\documentclass{article}

\PassOptionsToPackage{numbers}{natbib}
\usepackage{xspace}

\usepackage[preprint]{neurips_2026}

\usepackage[utf8]{inputenc}
\usepackage[T1]{fontenc}
\usepackage{hyperref}
\usepackage{url}
\usepackage{booktabs}
\usepackage{amsfonts}
\usepackage{amsmath}
\usepackage{amssymb}
\usepackage{microtype}
\usepackage{xcolor}
\usepackage{graphicx}
\usepackage{subcaption}
\usepackage{multirow}
\usepackage{makecell}
\usepackage{enumitem}
\usepackage{cleveref}
\usepackage{tcolorbox}
\usepackage{longtable}
\usepackage{array}
\usepackage{colortbl}
\usepackage{wrapfig}
\usepackage{parskip}
\usepackage{placeins}

\tcbuselibrary{skins}
\definecolor{darkgreen}{rgb}{0.0, 0.5, 0.0}
\newcommand{\datasetname}{\textsc{SpatialChain}\xspace}

\title{SpatialChain: A Benchmark for Auditing Spatial Reasoning Faithfulness in VLMs}

\author{%
  Rafael Teixeira Sousa\textsuperscript{1,2,3} \quad
  Vinícius Paulo Lopes de Oliveira\textsuperscript{2,3} \quad
  Elisa Ayumi Masasi de Oliveira\textsuperscript{2,3} \\
  Luiza Martins de Freitas Cintra\textsuperscript{2,3} \quad
  Fernanda Bufon Färber\textsuperscript{2,3} \quad
  Igor Dias Aguiar\textsuperscript{2,3} \\
  Julia Yasmim de Almeida Nobre\textsuperscript{2,3} \quad
  Arlindo Rodrigues Galvão Filho\textsuperscript{2,3} \\[6pt]
  \textsuperscript{1}Universidade Federal de Mato Grosso (UFMT) \quad
  \textsuperscript{2}Universidade Federal de Goiás (UFG) \\[2pt]
  \textsuperscript{3}Advanced Knowledge Center for Immersive Technologies (AKCIT) \\[2pt]
  \texttt{rafael.sousa@ufmt.br}
}

\begin{document}

\maketitle
\begin{abstract}
Thinking-enabled vision-language models (VLMs) report ever-higher accuracy on spatial benchmarks, yet final-answer scores cannot reveal whether a correct prediction reflects faithful spatial reasoning or a linguistic shortcut.
We introduce \datasetname, a dataset of 28{,}350 training and 899 test examples pairing spatially-oriented GQA questions with scene-graph-grounded reasoning chains — retained only when the generated answer matches the symbolic ground truth — and a two-axis evaluation combining objective chain-overlap metrics with a scene-graph-aware LLM judge that scores faithfulness and completeness independently of the final answer.
Applied to nine thinking-enabled VLMs, the protocol surfaces three findings invisible to standard accuracy: (i)~four of nine models achieve $\geq\!79\%$ VQA accuracy while exhibiting shortcut rates above $39\%$ — correct answers whose reasoning the judge marks as unfaithful; (ii)~chain quality significantly predicts answer correctness for seven of nine models, but the two exceptions (Claude~Sonnet~4.6, InternVL3.5-8B) reveal qualitatively distinct failure modes — terse output vs.\ verbose-decorative reasoning — that benchmark accuracy alone conflates; (iii)~SFT on \datasetname{} improves Qwen3-VL-8B by $+6.2$\,pp in-domain and reduces its shortcut rate to $22\%$, while a possible stylistic specialization effect on external benchmarks motivates replay-augmented training as mitigation.
The faithfulness judge is validated against 198 human-annotated items, where judge--human agreement matches human--human agreement, and against a second judge from a different provider, which preserves the model ranking ($\rho=0.88$).
Data, generation scripts, and evaluation code are released at \url{github.com/spatialchain/SpatialChainBenchmark}.
\end{abstract}

\section{Introduction}
\label{sec:intro}

Vision-language models (VLMs) have achieved remarkable performance on visual question answering (VQA) and spatial understanding benchmarks~\citep{yue2024mmmu, Lu_2025_CVPR}. More recently, several model families have been augmented with explicit multimodal reasoning, including Qwen3-VL~\citep{bai2025qwen3}, InternVL3~\citep{wang2025internvl3}, GLM-4.1-9B~\citep{glm2025}, Claude~4.6~\citep{anthropic2026claude46}, and GPT-5.4~\citep{openai2026gpt54}. These thinking-enabled models drive substantial gains on challenging benchmarks~\citep{yue2025mmmu, wang2024measuring, du2024embspatial}, yet high accuracy alone does not establish that a model is \emph{reasoning} rather than exploiting statistical language priors. A model that has memorized common object co-occurrences (e.g., \emph{cup on table}, \emph{person on chair}) may answer spatial queries correctly without performing any actual spatial inference. Standard benchmarks provide no mechanism to distinguish these two regimes: only the final answer is evaluated, leaving the underlying reasoning process entirely unaudited~\citep{goyal2017making, agrawal2018vqa, ma2024robust, de2023visual}.

To address this, we present \textbf{\datasetname}, which attaches an explicit, graph-grounded reasoning chain to each spatially-oriented question drawn from GQA~\citep{hudson2019gqa}. Every chain is synthesized from GQA's structured scene graphs via Claude~Haiku~4.5 with extended thinking, retained only when the generated answer matches the ground-truth answer, and stored in a format compatible with thinking-mode VLMs.
Because each reasoning step derives from a verifiable symbolic annotation, \datasetname{} enables a new evaluation regime: beyond \emph{did the model answer correctly}, we can ask \emph{did its reasoning trace engage with the correct spatial relations} — and detect shortcut behavior even when the final answer is right.

\paragraph{Contributions.}
\begin{itemize}[leftmargin=*, noitemsep]
  \item We introduce \datasetname, a dataset of 28{,}350 training and
        899 test examples pairing spatially-oriented GQA questions with
        scene-graph-grounded reasoning chains, providing the first
        structured reference against which the relational content
        of a VLM's thinking trace can be audited.
  \item We propose a two-axis evaluation protocol combining objective
        chain-overlap metrics with a scene-graph-aware LLM judge that
        scores reasoning faithfulness and completeness independently of
        the final answer, and show across nine thinking-enabled VLMs that
        chain quality significantly predicts answer correctness for seven
        of nine models — while exposing shortcut rates above 39\% in four
        models that achieve $\geq 79\%$ VQA accuracy.
        We validate the judge against human annotators, a second judge
        from a different provider, and a sweep of the faithfulness
        threshold (Appendix~\ref{app:judge-validation}).
  \item We show that SFT on \datasetname{} improves Qwen3-VL-8B by
        $+6.2$\,pp in-domain and reduces its shortcut rate to 22\% —
        the lowest of all evaluated models — while characterizing a
        stylistic specialization effect on out-of-distribution spatial
        benchmarks and proposing replay-augmented training as mitigation.
\end{itemize}
\noindent Data, generation scripts, and evaluation code are released
at \url{github.com/spatialchain/SpatialChainBenchmark}.

\section{Related Work}
\label{sec:related}

Thinking-enabled VLMs have driven substantial gains on general VQA suites~\citep{yue2024mmmu, yue2025mmmu} and on dedicated spatial benchmarks such as EmbSpatial-Bench~\citep{du2024embspatial}, ERQA~\citep{team2025gemini}, VSR~\citep{liu2023visual}, and VSI-Bench~\citep{yang2025thinking}; SpatialRGPT~\citep{cheng2024spatialrgpt} further improves spatial perception by grounding VLMs in 3D scene graphs, while recent probes expose persistent failures on geometric consistency and subtle spatial comparisons~\citep{khangaonkar2026spatial, kim2026vlmsubtlebench}. All of these benchmarks share the same structural limitation: they score only the final answer, providing no mechanism to determine whether a correct prediction reflects genuine spatial inference or a linguistic shortcut~\citep{goyal2017making, agrawal2018vqa}. GQA~\citep{hudson2019gqa} exposes relational structure through scene graphs that could in principle support process-level evaluation, but its standard protocol is equally answer-only.

Inspired by chain-of-thought prompting~\citep{wei2022chain}, several efforts expose the reasoning process alongside the final answer. ScienceQA~\citep{lu2022learn} and VCR~\citep{zellers2019recognition} provide human rationales in specialised domains, with free-form text that lacks structured grounding to image content. A more targeted line associates reasoning steps with image regions — Visual-CoT~\citep{shao2024visual}, CogCoM~\citep{qi2024cogcom}, DualFocus~\citep{cao2024dualfocus}, Chain-of-Spot~\citep{liu2024chain}, and, at scale, VisReason~\citep{li2025visreason} — and EmbodiedVSR~\citep{zhang2025embodiedvsr} uses dynamic scene graphs to guide chain-of-thought on spatial tasks. Closer to our evaluation goal, SpatialEval~\citep{wang2024spatialeval} scores final answers on spatial tasks without auditing the trace, and MME-CoT~\citep{jiang2025mmecot} evaluates reasoning quality against free-form or human-written rationales without structured grounding to image content. These resources advance \emph{visual} grounding — \emph{where to look} — but none provides a verifiable reference for the \emph{relational content} of a chain: a model that crops the correct region but reasons about the wrong spatial relation remains indistinguishable from one that genuinely understands the scene. \datasetname{} fills this gap by deriving every reasoning step from GQA's structured scene graphs and retaining chains only when consistent with the symbolic ground truth, enabling an evaluation that asks not only \emph{did the model look at the right region} but \emph{did it reason about the correct spatial relations}.

\section{The \datasetname{} Dataset}
\label{sec:dataset}

\subsection{Source Data: GQA}
\label{sec:dataset:gqa}

GQA~\citep{hudson2019gqa} provides 113{,}018 images from Visual
Genome~\citep{krishna2017visual} with structured scene graphs encoding
objects, attributes, and pairwise spatial relations (e.g.,
\texttt{left of}, \texttt{behind}, \texttt{above}), alongside
question--answer pairs annotated with a rich type taxonomy.
We select exclusively questions that require resolving at least one
spatial relation, following a two-stage pipeline.

\subsection{Question Selection Pipeline}
\label{sec:dataset:selection}

\paragraph{Stage 1 — Type-based filtering.}
We retain questions whose GQA type falls into one of 17
relation-centric categories that structurally require spatial or
relational reasoning (full list in Appendix~\ref{app:gqa-types}).
Applied to the GQA balanced training split, this yields 135{,}061
candidates from 49{,}503 unique images.
For the test split, we additionally rank candidates by the length of
their GQA \texttt{semanticStr} field — a proxy for relational chain
complexity — and retain only the 1{,}000 longest, biasing the test
set toward compositional, multi-step spatial queries.

\paragraph{Stage 2 — LLM spatial verification.}
Type-based filtering is necessary but not sufficient: many
relation-type questions can be answered by object recognition alone
(e.g., \textit{``Is the dog sitting?''} uses a \texttt{relVerify}
template but requires no spatial inference).
We therefore apply a second filter using GPT-5-nano as a binary
classifier, distinguishing questions that require relative position,
proximity, containment, or directional reasoning (\textsc{yes}) from
those answerable by attribute recognition or existence detection
alone (\textsc{no}).
The prompt is reproduced in Appendix~\ref{app:spatial-filter-prompt}.

After both stages we obtain 86{,}773 candidate training
and 900 test candidate examples, each paired with the corresponding
scene-graph subgraph.
Due to the computational cost of extended-thinking generation,
we randomly sample 30{,}000 candidates for reasoning chain generation.
The remaining 56{,}773 filtered candidates carry no annotations
beyond the GQA-derived selection and are not used in any experiment
reported here; they may be released in a future version of \datasetname{}.

\subsection{Reasoning Chain Generation}
\label{sec:dataset:generation}

Chains are generated with Claude~Haiku~4.5 (extended thinking,
budget 3{,}000 tokens, max output 4{,}096 tokens) via the Anthropic
Message Batches API.
Each request provides the full scene-graph JSON as internal context
alongside the question; a system prompt instructs the model to reason
\emph{as if directly observing the image}, prohibiting any explicit
reference to scene-graph fields in its output.
The generated chain must therefore use natural visual language,
producing traces indistinguishable from those a human annotator would
write when looking at the image.
The full prompt is in Appendix~\ref{app:generation-prompt}.

\paragraph{Answer-consistent filtering and verifiable reference.}
The model's extended thinking trace is retained as the reasoning
chain only when its visible answer matches the GQA ground-truth
answer; chains accompanying incorrect answers are discarded.
A small number of additional examples are lost to generation
errors (malformed output or truncated traces),
bringing the final counts to \textbf{28{,}350 training} and
\textbf{899 test} examples.
Every retained chain therefore concludes correctly and was generated
conditioned on the scene graph, which makes it cross-checkable
against the symbolic annotations — the property that
distinguishes \datasetname{} from datasets where chains are
generated freely from image content without a verifiable
ground-truth structure~\citep{li2025visreason, shao2024visual},
and that underlies the objective reasoning metrics in
\S\ref{sec:eval:objective}.
The filter verifies the conclusion, not each intermediate step:
a chain could reach the right answer through a locally unsupported
step (see \S\ref{sec:limitations}).
Representative examples, contrasting single-hop questions (one
spatial relation) with multi-hop questions that chain two relations
across anchor objects, are shown in Appendix~\ref{app:examples}.

\paragraph{Training target format.}
Retained chains are stored as
\texttt{<think>}\,\textit{[trace]}\,\texttt{</think>}
\texttt{<answer>}\,\textit{[answer]}\,\texttt{</answer>},
compatible with thinking-mode VLMs at both training and inference
time, allowing the reasoning process to be audited independently of
the final answer.

\subsection{Dataset Statistics}
\label{sec:dataset:stats}
\datasetname{} comprises \textbf{28{,}350 training} and \textbf{899 test} question--reasoning--answer triplets drawn from 19{,}263 and 719 unique images respectively (Table~\ref{tab:dataset-stats}). The relation distribution is dominated by projective relations: \texttt{left\_of}/\texttt{right\_of} account for 62.7\,\% of training questions under our first-occurrence labelling scheme, where each question is tagged according to the first spatial relation appearing in its string. Thus, these statistics indicate the entry point of the spatial chain rather than its complete relational content, since many questions combine multiple relations, especially in the test split.
In the test split this dominance follows from the complexity-biased selection (\S\ref{sec:dataset:selection}) — GQA's longest chains are built from projective relations — so per-relation cells with $n<20$ are non-inferential.
Training chains have a mean length of 115 words (median 112), and test chains are slightly longer (mean 145 words), reflecting the complexity-biased test selection. Full relation and chain-length breakdowns are in Appendix~\ref{app:dataset-extra}.

\begin{table}[t]
  \centering
  \caption{%
    \datasetname{} splits.
    \emph{Yes\,/\,No\,/\,Attr.} partitions answers between binary
    spatial verification and attribute-typed answers.
    \emph{Avg.\ chain} is the mean reasoning trace length in words.
  }
  \label{tab:dataset-stats}
  \small
  \begin{tabular}{lcccccc}
    \toprule
    \textbf{Split} & \textbf{Source} & \textbf{Samples}
      & \textbf{Images} & \textbf{Yes\,/\,No\,/\,Attr.}
      & \textbf{Yes-rate} & \textbf{Avg.\ chain (w)} \\
    \midrule
    Train & GQA train & 28{,}350 & 19{,}263
      & 16{,}516\,/\,9{,}926\,/\,1{,}908 & 0.58 & 115 \\
    Test  & GQA val   &     899  &     719
      & 566\,/\,329\,/\,4               & 0.63 & 145 \\
    \bottomrule
  \end{tabular}
\end{table}

\section{Evaluation Framework}
\label{sec:eval}

We evaluate models along two complementary axes: \emph{downstream accuracy} on the VQA task, and \emph{reasoning quality} assessed through objective corpus metrics and an LLM-based judge. The former measures whether the model produces the correct answer; the latter examines whether it does so for the right reasons.

\subsection{Downstream VQA Accuracy}
\label{sec:eval:vqa}

We frame the task as binary spatial VQA: each question has a \textsc{Yes}/\textsc{No} ground-truth answer derived from GQA's scene graphs~\citep{hudson2019gqa}. Accuracy is computed as exact-match after lower-casing and stripping articles, consistent with the standard GQA evaluation protocol. We additionally report accuracy conditioned on answer polarity (\emph{Yes-acc} and \emph{No-acc}) to diagnose positive-answer bias, a known marker of language-prior exploitation in VQA models~\citep{goyal2017making, agrawal2018vqa}.

\subsection{Objective Reasoning Metrics}
\label{sec:eval:objective}

We compare each model's thinking trace against the ground-truth scene-graph reasoning chain using ROUGE-1 and ROUGE-L~\citep{lin2004rouge}, token-level F1~\citep{rajpurkar2016squad} (after lower-casing and stop-word removal), entity overlap (fraction of scene-graph object labels present in the predicted chain), and length ratio (predicted-to-reference chain length in tokens, a proxy for verbosity). All metrics are computed on all test examples regardless of answer correctness, and stratified by correctness to test whether higher-quality chains co-occur with correct answers.
These metrics measure \emph{surface} agreement only: multiple valid traces can realise the same relational path (e.g., \emph{boy} vs.\ \emph{man} for one node), and overlap with a Claude-Haiku-generated reference rewards stylistic similarity, which fine-tuned models inherit. No faithfulness conclusion in this paper rests on them.

\subsection{LLM-as-Judge Reasoning Evaluation}
\label{sec:eval:judge}

Objective metrics capture only surface-level agreement with the reference output and are insufficient to assess whether a candidate rationale is semantically grounded in the scene representation or whether it covers all inference steps required to justify the answer. We therefore adopt a rubric-based LLM-as-a-judge protocol, following prior work showing that strong LLM judges can approximate human evaluation when guided by explicit criteria, while remaining sensitive to prompt design and bias \cite{zheng2023judging}. In particular, our prompt is inspired by rubric-based evaluation frameworks such as G-Eval, which uses structured prompting and chain-of-thought to score outputs under custom criteria \cite{liu2023geval}. The judge receives the question, ground-truth answer, scene graph, candidate final answer, and candidate reasoning, and returns three normalized scores in $[0,1]$, computed by Claude~Sonnet~4.6~\citep{anthropic2026claude46}: answer correctness ($\mathit{AC}$), reasoning faithfulness ($\mathit{RF}$), and reasoning completeness ($\mathit{RC}$).
Because the judge scores against the symbolic scene graph rather than the reference text, it is invariant to the surface variation that penalises overlap metrics. We validate it against human annotators and against a second judge from a different provider (GPT-5.6) in Appendix~\ref{app:judge-validation}.

Separating faithfulness from correctness is important because chain-of-thought explanations may be plausible yet unfaithful to the model's actual inference process \cite{turpin2023unfaithful}. Likewise, completeness captures whether the rationale accounts for all key spatial constraints and relations needed to support the final answer. Since scene graphs provide a structured representation of entities and relations, they serve as a natural evidence source for this judgment procedure and as the basis for the ground-truth reasoning chains themselves\cite{johnson2015imagegraphs,teney2017graphstructured,zhang2025embodiedvsr}.

\section{Experiments}
\label{sec:experiments}

\subsection{Models}
\label{sec:exp:models}

We evaluate seven systems spanning open-weight and proprietary VLMs with thinking capability. Open-weight models are evaluated in thinking mode via vLLM~\citep{kwon2023vllm} with continuous batching on a single A100 GPU (80\,GB): Qwen3-VL-8B and Qwen3-VL-4B~\citep{bai2025qwen3}, InternVL3.5-8B and InternVL3.5-4B~\citep{wang2025internvl3}, and GLM-4.1-9B~\citep{glm2025}. Proprietary models are queried through their respective APIs: Claude~Sonnet~4.6~\citep{anthropic2026claude46} and GPT-5.4~\citep{openai2026gpt54}. We additionally report two fine-tuned variants: Qwen3-VL-8B~FT and Qwen3-VL-4B~FT, described below.

\paragraph{Fine-tuning.}
We apply LoRA~\citep{hu2022lora} ($r=16$, $\alpha=32$, RSLoRA) to all linear layers of Qwen3-VL-8B and Qwen3-VL-4B, training for 2 epochs with an effective batch size of 12 on a single A100 80\,GB GPU. The training target includes the full scene-graph reasoning chain followed by the binary answer, formatted as \texttt{<think>\,$\cdots$\,</think> <answer>\,$\cdots$\,</answer>}. Full hyperparameters are in Appendix~\ref{app:hyperparams}.

\subsection{Inference Setup}
\label{sec:exp:setup}

Open-weight models are served with vLLM ($\tau=0.6$, top-$p=0.95$,
top-$k=20$, max 16{,}384 tokens). Fine-tuned checkpoints are
evaluated after merging the LoRA adapter into base weights.
Proprietary models use default temperature and receive identical
prompts; InternVL3.5 requires an additional prefix to activate
thinking mode. All models share the same system prompt
(Appendix~\ref{app:inference-prompt}). Full inference scripts are in the code release.\footnote{\url{https://github.com/spatialchain/SpatialChainBenchmark}}

\section{Results}
\label{sec:results}

We evaluate nine systems on the \datasetname{} test split:
the base and fine-tuned variants of Qwen3-VL at 4B and 8B,
two open-source baselines (InternVL3.5-4B/8B, GLM-4.1-9B), and
two proprietary thinking-mode models (Claude~Sonnet~4.6, GPT-5.4).
All models receive the same system prompt and produce a reasoning
chain followed by a binary \textsc{Yes}/\textsc{No} answer.
Four of the 899 test examples have non-binary ground-truth answers
(\emph{Attr.} in Table~\ref{tab:dataset-stats}) and are excluded from
accuracy only ($n=895$); chain-quality and judge metrics use all 899.
Unparsed predictions are excluded per model from that model's accuracy
(Table~\ref{tab:model_comparison}, $\dagger$) and do not affect these counts.

\subsection{Main Results}
\label{sec:res:main}

Table~\ref{tab:model_comparison} reports VQA accuracy alongside
chain-quality metrics measured against the ground-truth
scene-graph chains.

\begin{table*}
  \centering
  \caption{%
    Model comparison on the \datasetname{} test split ($n=895$; see \S\ref{sec:results}).
    \emph{Acc} and \emph{F1} are overall accuracy and macro-F1 on the
    binary \textsc{Yes}/\textsc{No} task.
    \emph{Yes}/\emph{No} are accuracies conditional on ground-truth
    polarity.
    \emph{R-1}, \emph{R-L}, \emph{F1\textsubscript{tok}} are
    reasoning-chain overlap metrics against the scene-graph reference
    chain.
    \emph{Len$_\text{ratio}$} is predicted-to-reference chain length
    (reference average: 145 tokens).
    $^\dagger$ Unparsed predictions excluded from accuracy
    (InternVL3.5-4B: 25; InternVL3.5-8B: 16; Claude~Sonnet~4.6: 7).
  }
  \label{tab:model_comparison}
  \resizebox{0.9\linewidth}{!}{%
  \begin{tabular}{lcccccccc}
    \toprule
    Model
      & Acc & F1
      & Yes & No
      & R-1 & R-L & F1\textsubscript{tok}
      & Len$_\text{ratio}$ \\
    \midrule
    Qwen3-VL-8B FT
      & \textbf{85.03} & \textbf{88.68}
      & \textbf{92.76} & 71.73
      & 0.654 & 0.443 & 0.643 & 0.90 \\
    Qwen3-VL-4B FT
      & 82.23 & 86.67
      & 91.34 & 66.57
      & \textbf{0.657} & \textbf{0.450} & \textbf{0.646} & 0.89 \\
    \midrule
    GLM-4.1-9B
      & 81.34 & 84.49
      & 80.39 & \textbf{82.98}
      & 0.438 & 0.301 & 0.421 & 1.47 \\
    InternVL3.5-4B$^\dagger$
      & 81.34 & 87.43
      & 87.28 & 71.12
      & 0.314 & 0.227 & 0.305 & 3.01 \\
    Claude~Sonnet~4.6$^\dagger$
      & 79.89 & 85.15
      & 87.63 & 66.57
      & 0.359 & 0.258 & 0.354 & 0.36 \\
    InternVL3.5-8B$^\dagger$
      & 79.33 & 85.29
      & 86.57 & 66.87
      & 0.323 & 0.228 & 0.316 & 7.07 \\
    Qwen3-VL-8B
      & 78.88 & 82.83
      & 80.57 & 75.99
      & 0.425 & 0.293 & 0.412 & 4.06 \\
    Qwen3-VL-4B
      & 78.44 & 82.01
      & 77.74 & 79.64
      & 0.403 & 0.279 & 0.392 & 5.89 \\
    GPT-5.4
      & 72.18 & 78.37
      & 79.68 & 59.27
      & 0.296 & 0.225 & 0.294 & 0.25 \\
    \bottomrule
  \end{tabular}%
  }
\end{table*}

\paragraph{Fine-tuning yields the largest gains, but the field is
tight.}
Qwen3-VL-8B~FT leads with $85.0\%$ accuracy, a $+6.2$\,pp gain over
its base model.
GLM-4.1-9B and InternVL3.5-4B tie as the strongest zero-shot systems
at $81.3\%$, with Claude~Sonnet~4.6 close behind at $79.9\%$.
GPT-5.4 is the weakest model overall ($72.2\%$), trailing the 4B
Qwen3 base by 6\,pp despite being substantially larger.

\paragraph{Chain-quality metrics partition models into three regimes.}
Fine-tuned models produce chains closely matching the reference in
both length ($\text{ratio}\approx 0.9$) and content (R-1~$>0.65$).
Open-weight base models (Qwen3, InternVL3.5) generate chains
$3$--$7\times$ longer than the reference with moderate overlap.
Proprietary models (Claude~Sonnet~4.6, GPT-5.4) are at the opposite
extreme: chains under $0.4\times$ reference length, effectively
committing to a final answer with minimal explicit reasoning.
GLM-4.1-9B sits between the two regimes, generating chains of
near-reference length ($\text{ratio}=1.47$) with the highest R-1
among non-fine-tuned models.

\paragraph{Reasoning quality and answer accuracy are not aligned
across model families.}
The relationship between R-1 and accuracy is monotonic only within
the Qwen3 family.
Across families, GPT-5.4's R-1 of $0.30$ corresponds to $72\%$
accuracy, while InternVL3.5-4B's R-1 of $0.31$ corresponds to
$81\%$.
Claude~Sonnet~4.6 and InternVL3.5-8B sit at similar R-1 levels
($\approx 0.34$) yet differ in accuracy by less than 1\,pp despite a
$20\times$ difference in chain length.
Formally, across all nine models accuracy and R-1 correlate only
moderately (Pearson $r=+0.69$, $p=0.04$; Spearman $\rho=+0.64$,
$p=0.06$), and the association collapses once the two fine-tuned
variants are removed ($r=+0.45$, $p=0.32$; Table~\ref{tab:crossmodel-corr}).
This decoupling motivates the within-model analysis in
\S\ref{sec:res:correlation}.

\subsection{Yes/No Asymmetry as a Signature of Language Priors}
\label{sec:res:polarity}

Table~\ref{tab:model_comparison} reveals a striking polarity gap
$\Delta_\text{pol} = \text{Yes-acc} - \text{No-acc}$. The pattern is not driven by model scale or family alone: it
separates models that confirm spatial descriptions regardless of
visual evidence (yes-biased) from models whose reasoning admits
mismatch between question and image (balanced).
We interpret a large $\Delta_\text{pol}$ as a signature of
language-prior exploitation: binary questions of the form
\textit{``Is there X...''} are known to carry strong yes-biases
in natural image corpora~\citep{zhang2016yin, goyal2017making,
agrawal2018vqa}, and a model defaulting to confirmation
regardless of visual evidence would exhibit exactly this pattern.

\subsection{Does Reasoning Quality Predict Answer Correctness?}
\label{sec:res:correlation}

The averages in Table~\ref{tab:model_comparison} do not tell us
whether higher-quality chains are produced \emph{when the model is
right} — the central question \datasetname{} is designed to answer.
For each model, we partition the test set into correct and incorrect
predictions and compare R-1 and token-level F1 using a two-sided
Mann-Whitney U test (Table~\ref{tab:correlation}): a significant
positive $\Delta$ indicates that chain quality carries information
about spatial correctness beyond what the final answer alone reveals.

\begin{table}
  \centering
  \caption{Reasoning-chain quality conditioned on answer correctness.
    $\Delta$R-1 is the mean R-1 difference between correct and
    incorrect predictions.
    $^*$\,$p < 0.05$; $^{**}$\,$p < 0.01$.}
  \label{tab:correlation}
  \small
  \begin{tabular}{lcccccc}
    \toprule
    Model & R-1$_\checkmark$ & R-1$_\times$ & $\Delta$R-1
      & F1$_\checkmark$ & F1$_\times$ & $p$ \\
    \midrule
    Qwen3-VL-4B
      & 0.418 & 0.351 & \textbf{+0.067}
      & 0.406 & 0.342 & $<\!0.001^{**}$ \\
    Qwen3-VL-8B
      & 0.439 & 0.373 & +0.066
      & 0.426 & 0.362 & $<\!0.001^{**}$ \\
    GLM-4.1-9B
      & 0.447 & 0.399 & +0.048
      & 0.429 & 0.385 & $<\!0.001^{**}$ \\
    Qwen3-VL-4B FT
      & 0.664 & 0.625 & +0.039
      & 0.653 & 0.612 & $<\!0.001^{**}$ \\
    Qwen3-VL-8B FT
      & 0.660 & 0.626 & +0.034
      & 0.649 & 0.612 & $<\!0.001^{**}$ \\
    InternVL3.5-4B
      & 0.320 & 0.288 & +0.031
      & 0.311 & 0.282 & $0.003^{**}$ \\
    GPT-5.4
      & 0.297 & 0.292 & +0.004
      & 0.295 & 0.290 & $0.027^{*}$ \\
    \midrule
    Claude~Sonnet~4.6
      & 0.361 & 0.354 & +0.007
      & 0.356 & 0.347 & $0.071$ \\
    InternVL3.5-8B
      & 0.323 & 0.322 & +0.001
      & 0.317 & 0.316 & $0.458$ \\
    \bottomrule
  \end{tabular}
\end{table}

Seven of nine models show a significant positive correlation
($p < 0.05$): all Qwen3 variants, GLM-4.1-9B, InternVL3.5-4B, and
GPT-5.4.
The effect is strongest in the Qwen3 base models
($\Delta\text{R-1}\approx +0.067$), moderate in the fine-tuned models
($\approx +0.035$ — compressed by uniformly higher chain quality in
both groups), and weak but detectable in GPT-5.4 ($+0.004$, $p=0.03$).

The two exceptions expose qualitatively distinct failure modes.
Claude~Sonnet~4.6's non-significance ($p=0.07$, $\Delta=+0.007$) stems from
extremely short output ($\text{Len}_\text{ratio}=0.36$): there is
too little reasoning content for quality to vary meaningfully across
the two groups.
InternVL3.5-8B is the more revealing case: the longest chains in our
evaluation ($\text{Len}_\text{ratio}=7.07$) yet zero correlation
with correctness ($\Delta=+0.001$, $p=0.46$).
This is verbose output that \emph{resembles} reasoning but whose
content is decoupled from the spatial judgment being made —
a failure mode chain length alone cannot detect.

Taken together, the table illustrates \datasetname{}'s central
diagnostic value: GPT-5.4 ($72\%$ accuracy, significant correlation)
and Claude~Sonnet~4.6 ($80\%$ accuracy, non-significant) produce equally
compact chains yet differ fundamentally in whether their output
carries information about correctness — a distinction that benchmark
accuracy cannot surface.
This is consistent with evidence that intermediate tokens need not carry the reasoning behind a correct answer~\citep{stechly2025beyondsemantics}; \datasetname{} says \emph{which} systems are in that regime. Stratifying by hop count (Table~\ref{tab:hops}) shows multi-hop questions are not harder: the dominant difficulty is perceiving individual relations, not composing them.

\subsection{Cross-Benchmark Behaviour After Fine-Tuning}
\label{sec:res:tradeoff}
 
SFT on \datasetname{} improves Qwen3-VL-4B in-domain
($0.784 \to 0.822$, $+3.8$\,pp) but degrades it on both external
spatial benchmarks: ERQA $0.453 \to 0.380$ ($-7.3$\,pp) and
EmbSpatial-Bench $0.791 \to 0.757$ ($-3.4$\,pp).
Inspection of outputs reveals the mechanism: after fine-tuning,
$100\%$ of outputs adopt the fixed ``Looking at\ldots'' opener
(vs.\ $0\%$ before), and average chain length collapses to $15\%$ of
its pre-SFT value on ERQA.
The model has learned to produce \datasetname{}-style chains and
applies this format even when the input distribution differs —
stylistic specialization rather than loss of underlying spatial
knowledge, as evidenced by the non-monotonic trajectory across checkpoints and by the preservation of in-plane spatial relations on EmbSpatial-Bench.
Per-checkpoint curves and per-category breakdowns are reported in
Appendix~\ref{app:extra-results}.

\subsection{LLM-as-Judge Evaluation}
\label{sec:res:judge}

\begin{wrapfigure}{r}{0.6\linewidth}
  \vspace{-10pt}
  \centering
  \includegraphics[width=\linewidth]{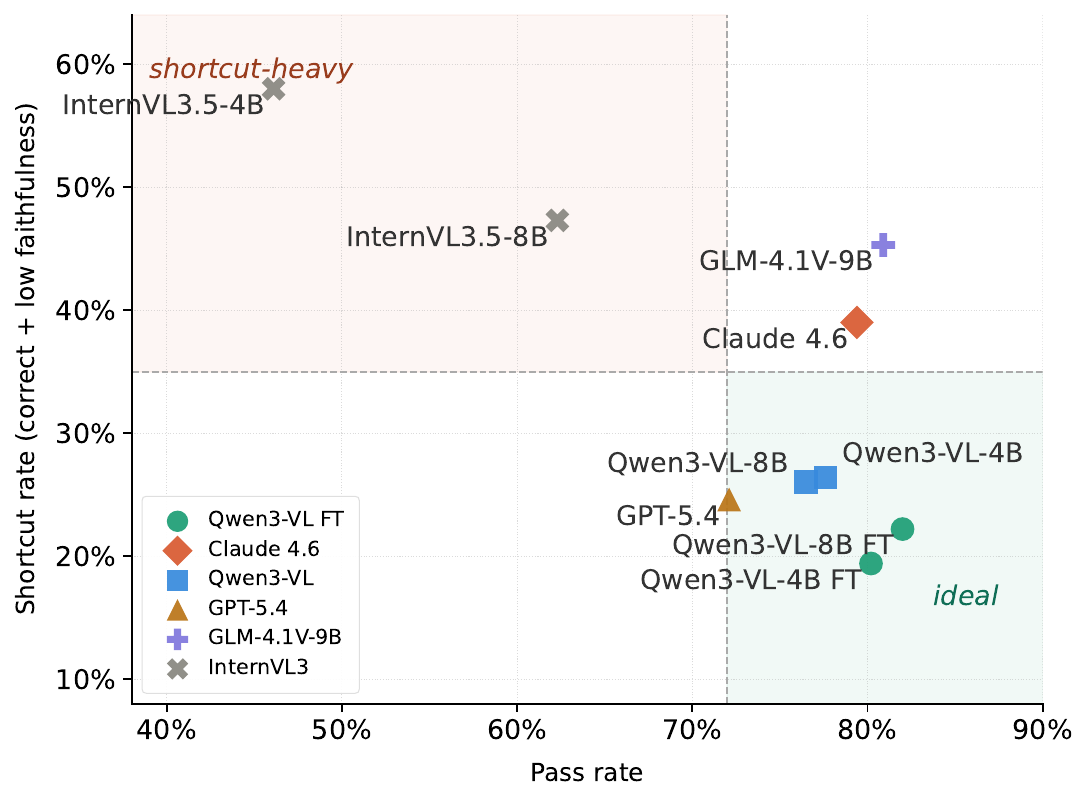}
  \caption{Pass rate vs.\ shortcut rate. The bottom-right region
    (\emph{ideal}) marks high passing with low shortcut behaviour;
    the top-left (\emph{shortcut-heavy}) marks competitive accuracy
    achieved with frequent unfaithful reasoning. Fine-tuned Qwen3-VL
    variants are closest to ideal; InternVL3.5-4B is the clearest
    shortcut case.}
  \label{fig:accuracy_vs_shortcut}
  \vspace{-8pt}
\end{wrapfigure}

Objective metrics measure surface overlap with the reference chain but cannot assess semantic faithfulness. We run the scene-graph-aware judge on all nine models (Table~\ref{tab:judge} and Figure~\ref{fig:accuracy_vs_shortcut}) and additionally report the \emph{shortcut rate}: the fraction of correct test items whose chain the judge marks as low-faithfulness ($\mathit{RF}\leq 0.5$), computed as in the released code. The inclusive cutoff is deliberate: a score of exactly $0.5$ marks a right conclusion that is weakly grounded (typically a chain that restates the question), which human annotators also score low (Appendix~\ref{app:judge-validation}).

\begin{table}
  \centering
  \caption{LLM-as-judge scores, sorted by pass rate.
    \emph{Faith.}\ and \emph{Compl.}\ are mean faithfulness and
    completeness in $[0,1]$.
    \emph{Pass} requires $\mathit{AC}\geq 0.5$ and
    $(\mathit{RF}+\mathit{RC})/2\geq 0.5$.
    \emph{Shortcut} is the fraction of correct test items with
    $\mathit{RF}\leq 0.5$.}
  \label{tab:judge}
  \small
  \begin{tabular}{lccccc}
    \toprule
    Model & Acc &  Faith. & Compl. & Pass & Shortcut \\
    \midrule
    Qwen3-VL-8B FT & \textbf{85.03} & 0.630 & 0.700 & \textbf{0.820} & 0.222 \\
    Qwen3-VL-4B FT & 82.23 & \textbf{0.631} & \textbf{0.708} & 0.802
      & \textbf{0.194} \\
    GLM-4.1-9B     & 81.34 & 0.527 & 0.583 & 0.809 & 0.453 \\
    Claude~Sonnet~4.6     & 79.89 & 0.541 & 0.582 & 0.794 & 0.390 \\
    Qwen3-VL-4B    & 78.44 & 0.585 & 0.658 & 0.776 & 0.264 \\
    Qwen3-VL-8B    & 78.88 & 0.571 & 0.658 & 0.765 & 0.260 \\
    GPT-5.4        & 72.18 & 0.542 & 0.568 & 0.721 & 0.246 \\
    InternVL3.5-8B & 79.33 & 0.443 & 0.488 & 0.623 & 0.473 \\
    InternVL3.5-4B & 81.34 & 0.346 & 0.360 & 0.461 & 0.580 \\
    \bottomrule
  \end{tabular}
\end{table}

\paragraph{Fine-tuned models produce the most faithful reasoning.}
Both Qwen3-VL FT variants lead on faithfulness ($\approx 0.63$) and completeness ($\approx 0.70$), with the lowest shortcut rates (19--22\%), corroborating the objective metrics: SFT on scene-graph chains teaches grounded spatial reasoning, not just answer matching.

\paragraph{High accuracy and high shortcut rate coexist.}
Four models achieve $\geq 79\%$ accuracy while exhibiting shortcut rates above $39\%$. Claude~Sonnet~4.6 ($79.9\%$, shortcut $39\%$) and InternVL3.5-4B ($81.3\%$, $58\%$) are the clearest cases: competitive benchmark scores mask reasoning that the judge flags as largely ungrounded. This finding directly extends the correlation analysis in \S\ref{sec:res:correlation}: chain quality is not only weakly correlated with correctness for these models, but the judge confirms that correct answers are frequently reached without faithful spatial inference.

\paragraph{Verbosity does not imply faithfulness.}
InternVL3.5-8B generates chains $7\times$ the reference length yet achieves the second-lowest pass rate ($62.3\%$) and a $47\%$ shortcut rate. The judge confirms quantitatively what was identified statistically before: long output that is decoratively verbose rather than spatially grounded.

\paragraph{High pass rate and high shortcut rate are not contradictory.}
GLM-4.1-9B achieves the third-highest pass rate ($80.9\%$) alongside a $45\%$ shortcut rate. This follows mechanically from the pass criterion averaging $\mathit{RF}$ and $\mathit{RC}$: a model can satisfy $(\mathit{RF}+\mathit{RC})/2\geq 0.5$ while many individual chains fall below the faithfulness cutoff. The two metrics answer different questions and should be read together, not reconciled.

\paragraph{The faithfulness signal is robust to the judge and the threshold.}
A second judge from a different provider (GPT-5.6, identical prompt) preserves the ranking of models by shortcut rate ($\rho=0.88$; faithfulness scores $\rho=0.74$, pass/fail $\kappa=0.84$), being uniformly stricter without changing the comparative conclusions. Shortcut rates are identical for $\tau\in[0.5,0.6]$, and the ordering holds wherever the metric separates models. On 198 human-annotated items, judge--human agreement on faithfulness ($\rho=0.49$) matches human--human agreement ($\rho=0.44$). Details in Appendix~\ref{app:judge-validation}.

\subsection{Qualitative Analysis}
\label{sec:analysis:qualitative}

Figure~\ref{fig:qualitative-example} (Appendix~\ref{app:quali-fig}) illustrates the two most contrasting reasoning regimes on a representative test example. The fine-tuned model traces the spatial chain step-by-step, closely mirroring the scene-graph reference. InternVL3.5-8B produces the correct answer, but buries it in hallucinated scene descriptions (\textit{``near a body of water''}), the verbal signature of its null $\Delta$R-1 and $47\%$ shortcut rate. The full comparison across all nine models, including the base model and GLM-4.1-9B's qualitatively distinct pattern, is in Appendix~\ref{app:extra-quali-samples}.

\section{Limitations}
\label{sec:limitations}

\datasetname{} has five main limitations.
First, GQA scene graphs were annotated by humans and may contain errors
that propagate into the synthetic reasoning chains.
Second, the answer-consistency filter verifies a chain's conclusion,
not each intermediate step: the generator (Claude~Haiku~4.5) can
hallucinate, so retained chains conclude correctly but are not
guaranteed step-correct; per-step verification against scene-graph
node IDs is the natural next step.
Third, the faithfulness judge (Claude~Sonnet~4.6) is a single model
from the same family as the reference generator, and one evaluated
system shares that family. The second-judge and human-validation
results (Appendix~\ref{app:judge-validation}) bound this risk, but
references generated by a non-Claude model remain untested.
Fourth, the task is binary \textsc{Yes}/\textsc{No} with a $58\%$
yes-rate in training, which enables $\Delta_\text{pol}$ but caps the
accuracy axis and is amplified by SFT; extending beyond binary VQA is
the focus of future versions.
Fifth, scene-graph relations are symbolic rather than metric, so the
benchmark does not capture full geometric understanding (e.g., exact
distances or 3D depth), and the complexity-biased test split is
dominated by projective \texttt{left}/\texttt{right} relations.

\section{Conclusion}
\label{sec:conclusion}

We introduced \datasetname, a dataset pairing spatially grounded
GQA questions with scene-graph-consistent reasoning chains,
enabling training and structured evaluation of spatial reasoning
in thinking-enabled VLMs.
Chain quality significantly predicts answer correctness across most
models; the judge evaluation reveals that high accuracy frequently
coexists with shortcut rates above 39\% — evidence that answer
scores alone are insufficient to characterize spatial reasoning
ability — and this signal survives a change of judge, a sweep of
the threshold, and comparison with human annotators.
SFT improves in-domain accuracy while exposing a stylistic
trade-off on external benchmarks, motivating replay-augmented
training as a future direction.
Beyond SFT, \datasetname{} supports reinforcement learning —
outcome and process rewards can be computed automatically from
the verifiable answer and scene-graph reference — and provides
structured spatial supervision for Vision-Language-Action
models~\citep{qu2025spatialvla, zhang2025inspire}, where object-relation
understanding is a prerequisite for manipulation and navigation.
We hope \datasetname{} encourages the community to interrogate
the \emph{reasoning process} behind model predictions, beyond
the final answer.


\bibliographystyle{plainnat}
\bibliography{references.bib}

\appendix

\appendix

\section{GQA Relation-Centric Question Types}
\label{app:gqa-types}

Table~\ref{tab:gqa-types} lists the 17 GQA question types retained
in Stage~1 of the selection pipeline (\S\ref{sec:dataset:selection}),
grouped by the relational competence they probe.

\begin{table}[h]
  \centering
  \caption{GQA question types selected for \datasetname{}.}
  \label{tab:gqa-types}
  \small
  \begin{tabular}{ll}
    \toprule
    \textbf{Group} & \textbf{GQA types} \\
    \midrule
    Spatial relation verification
      & \texttt{relVerify}, \texttt{relVerifyCop}, \texttt{relVerifyCr} \\
    Existential with relation
      & \texttt{existRelS}, \texttt{existRelSC}, \texttt{existRelSRC} \\
    Place verification
      & \texttt{placeVerify}, \texttt{placeVerifyC} \\
    Two-object comparison
      & \texttt{twoDifferent}, \texttt{twoDifferentC},
        \texttt{twoSame}, \texttt{twoSameC} \\
    Category comparison
      & \texttt{diffAnimals}, \texttt{diffAnimalsC}, \texttt{diffGender} \\
    Attribute / ordinal selection
      & \texttt{chooseAttr}, \texttt{compare} \\
    \bottomrule
  \end{tabular}
\end{table}

\section{Spatial Filter Prompt}
\label{app:spatial-filter-prompt}

The following prompt is used with GPT-5-nano in Stage~2 of the
question selection pipeline to distinguish questions that genuinely
require spatial inference from those answerable by attribute
recognition or existence detection alone.

\begin{tcolorbox}[
  colback=gray!4, colframe=gray!35,
  title={Spatial Reasoning Filter — User Prompt},
  fonttitle=\small\bfseries,
  left=6pt, right=6pt, top=4pt, bottom=4pt
]
\small\ttfamily
You are building a benchmark to evaluate spatial reasoning in multimodal
LLMs.

Your task is to classify whether a question genuinely requires
\textbf{spatial reasoning} to be answered --- meaning the model must
understand the position, layout, distance, or relationship between
objects in an image.

\medskip
\textbf{Answer YES only if the question requires:}
\begin{itemize}[leftmargin=1.5em, noitemsep, topsep=2pt]
  \item Relative position between objects (left/right, above/below,
        in front/behind, next to, between)
  \item Counting objects within a spatial context (e.g., ``how many X
        are to the left of Y'')
  \item Distance or proximity judgment (e.g., ``what is closest to X'')
  \item Spatial containment or grouping (e.g., ``what is inside/outside X'')
  \item Layout understanding (e.g., ``what is in the corner/center/edge'')
  \item Sequential or directional arrangement
\end{itemize}

\medskip
\textbf{Answer NO if the question can be answered by:}
\begin{itemize}[leftmargin=1.5em, noitemsep, topsep=2pt]
  \item Simply detecting or recognizing an object (``is there a dog?'',
        ``what color is the car?'')
  \item Reading text or identifying attributes without spatial context
  \item Generic scene understanding without positional reasoning
        (``what is the weather?'')
  \item Yes/no questions about object existence
\end{itemize}

\medskip
\textbf{Examples:}\\
Q: What is to the left of the table? A: The chair $\rightarrow$ YES\\
Q: Is the dog sitting? A: Yes $\rightarrow$ NO\\
Q: What is between the lamp and the window? A: A plant $\rightarrow$ YES\\
Q: What color is the sky? A: Blue $\rightarrow$ NO\\
Q: How many people are behind the counter? A: Two $\rightarrow$ YES\\
Q: Is there a car in the image? A: Yes $\rightarrow$ NO

\medskip
Now classify the following:\\
Q: \{question\}\\
A: \{answer\}

\medskip
Reply with only YES or NO.
\end{tcolorbox}

\newpage
\section{Reasoning Chain Generation Prompt}
\label{app:generation-prompt}

The following system and user prompts are used to generate reasoning
chains with Claude~Haiku~4.5 (extended thinking enabled).

\begin{tcolorbox}[
  colback=gray!4, colframe=gray!35,
  title={Generation System Prompt},
  fonttitle=\small\bfseries,
  left=6pt, right=6pt, top=4pt, bottom=4pt
]
\small\ttfamily
You are given a structured description of an image (a scene graph with
objects, attributes, positions, and spatial relationships).
Use this information to understand the scene, but respond AS IF you were
directly looking at the image.

\medskip
IMPORTANT RULES FOR YOUR RESPONSE:
\begin{itemize}[leftmargin=1.5em, noitemsep, topsep=2pt]
  \item Never mention the scene graph, object IDs, coordinates, pixel
        positions, or any technical data.
  \item Never say things like ``according to the graph'',
        ``object X is at position (a, b)'', or reference any ID numbers.
  \item Write your reasoning as natural visual observation, as if you are
        describing what you see in a photograph.
        Use language like ``I can see...'', ``Looking at the image...'',
        ``The cheese appears to be...'', ``On the left side of the image...''.
  \item Describe spatial relationships naturally
        (e.g., ``the cheese is sitting to the left of the main dish''
        instead of ``cheese at x=1 is left of food at x=275'').
  \item Your response will be paired with the actual image later,
        so it must read as authentic visual reasoning.
\end{itemize}
\end{tcolorbox}

\begin{tcolorbox}[
  colback=gray!4, colframe=gray!35,
  title={Generation User Prompt Template},
  fonttitle=\small\bfseries,
  left=6pt, right=6pt, top=4pt, bottom=4pt
]
\small\ttfamily
Scene information (for your internal use only, do NOT reference directly):

\{scene\_graph\_text\}

\medskip
Question: \{question\}

\medskip
Think step-by-step about what you observe in the image,
then provide your final answer in the format:\\
ANSWER: $\langle$your answer$\rangle$
\end{tcolorbox}

\paragraph{Scene graph format.}
The \texttt{\{scene\_graph\_text\}} field is rendered as a JSON with structured
list of objects with the following fields per entry:
object name and ID, attributes (comma-separated), bounding box
$(x, y, w, h)$ in pixels, and outgoing spatial relations in the form
\texttt{relation\_name target\_name(target\_id)}.

\newpage
\section{Extended Dataset Statistics}
\label{app:dataset-extra}

\begin{figure}[h]
  \centering
  \includegraphics[width=\linewidth]{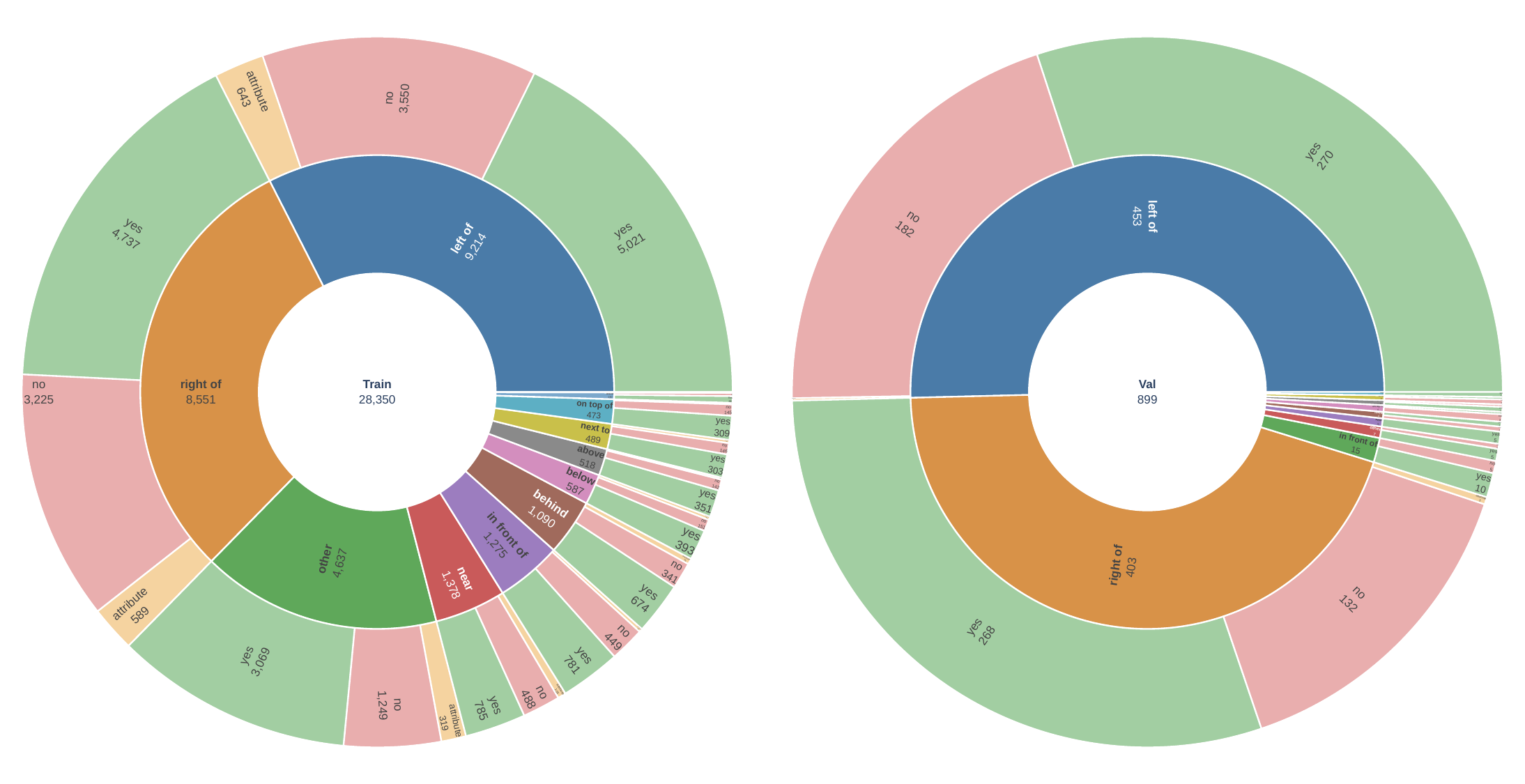}
  \caption{%
    Distribution of \datasetname{} training (left) and test (right)
    splits by primary spatial relation (inner ring) and answer class
    (outer ring; \textcolor{green!50!black}{green}: \textsc{yes},
    \textcolor{red!60!black}{red}: \textsc{no},
    \textcolor{orange!70!black}{orange}: attribute).
    The test split concentrates on projective relations, reflecting
    the complexity-biased selection.
  }
  \label{fig:sunburst-splits}
\end{figure}

\paragraph{Spatial relation distribution.}
Table~\ref{tab:relation-dist} reports the full breakdown of the
eleven spatial relation categories covered in \datasetname{},
together with per-relation yes-rates.
Projective relations (\texttt{left\_of}, \texttt{right\_of})
dominate both splits, consistent with GQA's underlying annotation
distribution.
The yes-rate varies substantially across relation types: vertical
relations (\texttt{above}, \texttt{on\_top\_of}) tend toward higher
yes-rates ($\approx 0.70$), while depth relations
(\texttt{in\_front\_of}, \texttt{behind}) are more balanced
($\approx 0.55$), reflecting the greater ambiguity of depth ordering
in 2D images. Mode details can be seen in Figure~\ref{fig:sunburst-splits}.

\begin{table}[h]
  \centering
  \caption{Per-relation statistics in the \datasetname{} training
    split. \emph{Yes-rate} is the fraction of questions with a
    \textsc{Yes} ground-truth answer.}
  \label{tab:relation-dist}
  \small
  \begin{tabular}{lrrr}
    \toprule
    \textbf{Relation} & \textbf{Count} & \textbf{\%} & \textbf{Yes-rate} \\
    \midrule
    \texttt{left\_of}    & 9{,}832 & 34.7 & 0.57 \\
    \texttt{right\_of}   & 7{,}919 & 27.9 & 0.58 \\
    \texttt{in\_front\_of} & 1{,}987 &  7.0 & 0.54 \\
    \texttt{behind}      & 1{,}406 &  5.0 & 0.55 \\
    \texttt{near}        &   347 &  1.2 & 0.61 \\
    \texttt{above}       &   763 &  2.7 & 0.71 \\
    \texttt{below}       &   412 &  1.5 & 0.68 \\
    \texttt{on\_top\_of} &   411 &  1.4 & 0.70 \\
    \texttt{next\_to}    & 1{,}876 &  6.6 & 0.59 \\
    \texttt{inside}      & 1{,}540 &  5.4 & 0.60 \\
    \texttt{between / other} & 1{,}857 &  6.6 & 0.52 \\
    \midrule
    \textbf{Total}       & 28{,}350 & 100.0 & 0.58 \\
    \bottomrule
  \end{tabular}
\end{table}

\begin{figure}
  \centering
  \includegraphics[width=\linewidth]{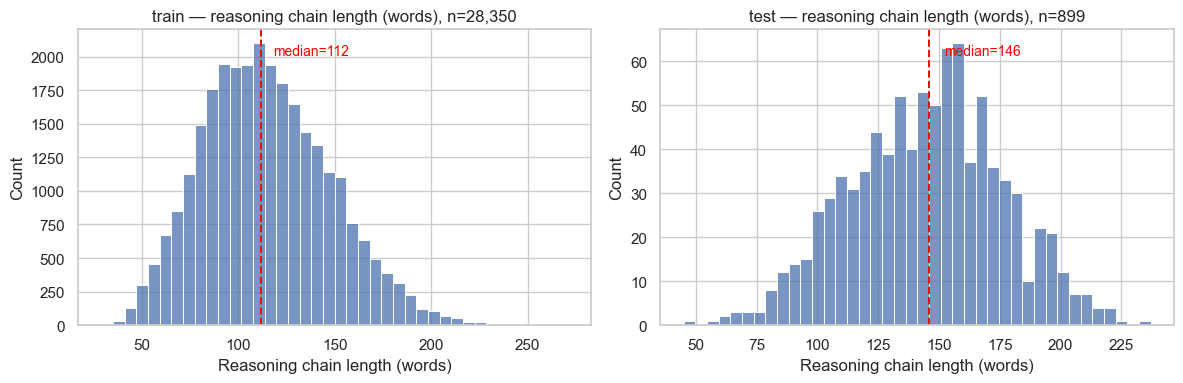}
  \caption{Word-count distribution of reasoning chains in the
    training (blue) and test (orange) splits.
    Dashed vertical lines indicate medians.}
  \label{fig:chain-length}
\end{figure}

\begin{figure}
  \centering
  \includegraphics[width=\linewidth]{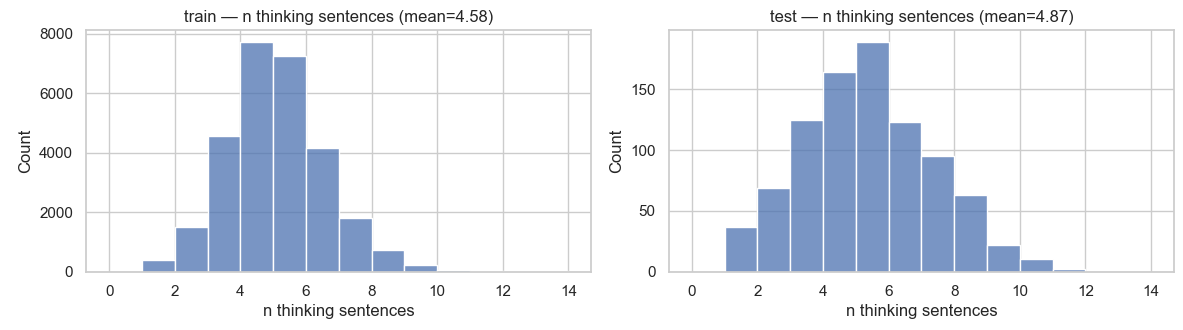}
  \caption{Distribution of the number of sentences per thinking trace for the training (left) and test (right) splits. Both distributions are unimodal and right-skewed, peaking at 5–6 sentences. The test split exhibits a slightly higher mean (4.87 vs. 4.58), consistent with its bias toward more complex multi-hop chains.}
  \label{fig:thinking_sentences}
\end{figure}

\paragraph{Reasoning chain length.}
Figure~\ref{fig:chain-length} shows the full word-count distributions
for training and test chains.
Both are unimodal and approximately log-normal.
Training chains span $P_{10}$--$P_{90}$ = [74, 159] words
(median 112, mean 115); test chains span [96, 203] words
(median 146, mean 145), reflecting the complexity-biased test
selection.
Fewer than 2\,\% of chains in either split exceed 250 words,
consistent with Claude~Haiku~4.5's tendency to produce concise,
focused reasoning under the natural-language grounding constraint.
At the sentence level (Figure~\ref{fig:thinking_sentences}), both
splits follow a right-skewed unimodal distribution centred around
5--6 sentences (train mean = 4.58; test mean = 4.87).
The slight rightward shift in the test split is consistent with the
complexity-biased selection: chains that resolve longer
spatial-relation sequences naturally require a few additional
reasoning steps.
Traces rarely exceed 10 sentences, confirming that the model's
chain-of-thought remains compact even for multi-hop queries.

\begin{figure}
  \centering
  \includegraphics[width=\linewidth]{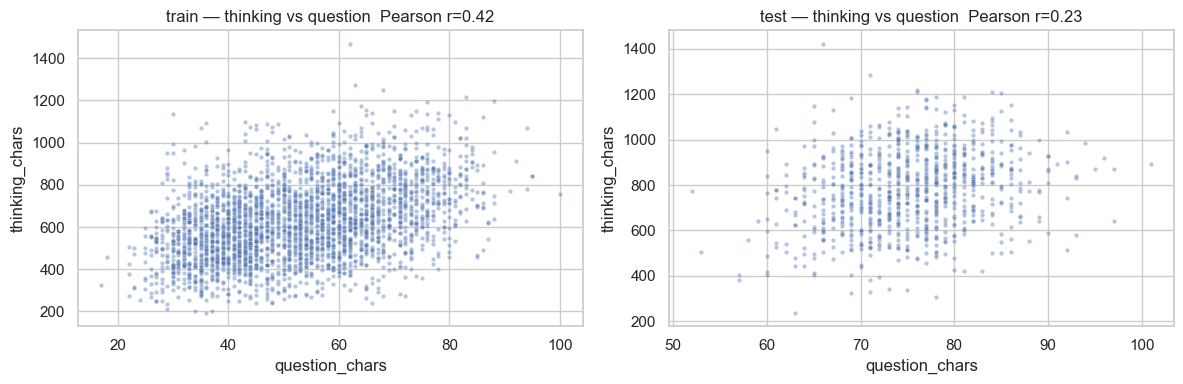}
  \caption{Thinking trace length (in characters) as a function of question length (in characters) for the training (left, Pearson r = 0.42) and test (right, Pearson r = 0.23) splits. Longer questions elicit longer reasoning traces.}
  \label{fig:thinking vs question}
\end{figure}

\paragraph{Thinking length vs. question length.}
Figure~\ref{fig:thinking vs question} plots thinking trace length (in characters) against question length for both splits. A moderate positive correlation is observed in the training set (Pearson r = 0.42), indicating that longer or more detailed questions tend to elicit proportionally longer reasoning traces. This correlation attenuates in the test split (r = 0.23), likely because test questions are more uniformly complex—compressing the range of question lengths while still requiring extended reasoning—which reduces the linear dependence between the two variables.

\section{Dataset Examples}
\label{app:examples}
\begin{figure*}[h!]
\centering
\small
\setlength{\fboxsep}{4pt}
\begin{minipage}[]{0.48\linewidth}
\begin{tcolorbox}[
  colback=green!4, colframe=green!45,
  title={\textbf{Single-hop · \textsc{Yes}} \hfill
         \textcolor{gray}{\texttt{img:2347569}}},
  fonttitle=\small\bfseries,
  left=4pt, right=4pt, top=3pt, bottom=3pt
]
  \centering
  \includegraphics[width=\linewidth, height=2.5cm,
                   keepaspectratio=True, clip]{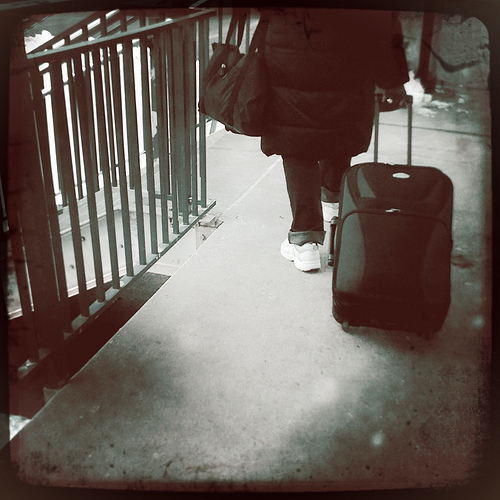}\\[4pt]
  \textit{Is there a fence \underline{to the left of} the
  person that is pulling a suitcase?''}\\[3pt]
  \textcolor{gray}{\scriptsize
  Looking at the image, I can see a person standing in the
  center-right area. [\ldots] To the \textbf{left} of this
  person, I can clearly observe a black fence running along
  the left side [\ldots]}\\[3pt]
  \textbf{Answer:} \textcolor{darkgreen}{\checkmark}~Yes,
  there is a fence to the left of the person.
\end{tcolorbox}
\end{minipage}
\hfill
\begin{minipage}[]{0.48\linewidth}
\begin{tcolorbox}[
  colback=red!4, colframe=red!40,
  title={\textbf{Single-hop · \textsc{No}} \hfill
         \textcolor{gray}{\texttt{img:2359192}}},
  fonttitle=\small\bfseries,
  left=4pt, right=4pt, top=3pt, bottom=3pt
]
  \centering
  \includegraphics[width=\linewidth, height=2.5cm,
                   keepaspectratio=True, clip]{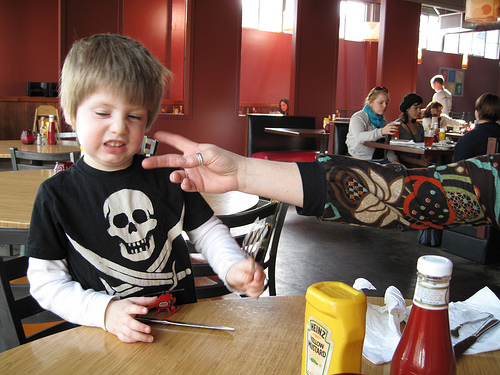}\\[4pt]
  \textit{Do you see any skateboards \underline{to the right of}
  the person that is wearing a shirt?''}\\[3pt]
  \textcolor{gray}{\scriptsize
  I can identify a boy on the left side wearing a black shirt.
  To the \textbf{right} of this person I can see red cans,
  napkins, silverware [\ldots] I do not see any skateboards
  in this area [\ldots]}\\[3pt]
  \textbf{Answer:} \textcolor{red!70!black}{$\times$}~No,
  there is a can to the right of the boy.
\end{tcolorbox}
\end{minipage}
\vspace{6pt}
\begin{minipage}[]{0.48\linewidth}
\begin{tcolorbox}[
  colback=green!4, colframe=green!45,
  title={\textbf{Multi-hop · \textsc{Yes}} \hfill
         \textcolor{gray}{\texttt{img:2325221}}},
  fonttitle=\small\bfseries,
  left=4pt, right=4pt, top=3pt, bottom=3pt
]
  \centering
  \includegraphics[width=\linewidth, height=2.5cm,
                   keepaspectratio=True, clip]{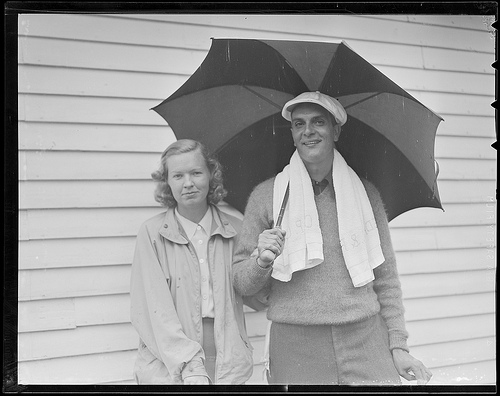}\\[4pt]
  \textit{Are there any umbrellas \underline{above} the people
  \underline{to the left of} the towel?''}\\[3pt]
  \textcolor{gray}{\scriptsize
  \textbf{Step 1 — left:}
  I can identify the towel — a white cloth on the right side
  of the image. To its \textbf{left} I can see a person
  standing.
  \textbf{Step 2 — above:}
  Examining what is \textbf{above} this person, I can clearly
  see umbrellas positioned overhead [\ldots]}\\[3pt]
  \textbf{Answer:} \textcolor{darkgreen}{\checkmark}~Yes,
  there is an umbrella above the people.
\end{tcolorbox}
\end{minipage}
\hfill
\begin{minipage}[]{0.48\linewidth}
\begin{tcolorbox}[
  colback=red!4, colframe=red!40,
  title={\textbf{Multi-hop · \textsc{No}} \hfill
         \textcolor{gray}{\texttt{img:2416648}}},
  fonttitle=\small\bfseries,
  left=4pt, right=4pt, top=3pt, bottom=3pt
]
  \centering
  \includegraphics[width=\linewidth, height=2.5cm,
                   keepaspectratio=True, clip]{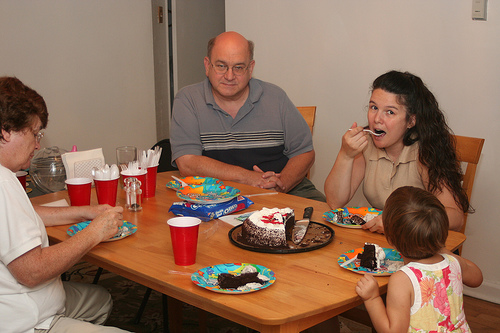}\\[4pt]
  \textit{Do you see pastries \underline{to the left of}
  the food that is \underline{to the left of} the knife?''}\\[3pt]
  \textcolor{gray}{\scriptsize
  \textbf{Step 1 — left:}
  Looking at the image, I can see a knife in the middle;
  to its \textbf{left} there are cookies on the table.
  \textbf{Step 2 — left:}
  To the \textbf{left} of the cookies I can see a woman,
  not pastries [\ldots]}\\[3pt]
  \textbf{Answer:} \textcolor{red!70!black}{$\times$}~No,
  there is a woman to the left of the cookies.
\end{tcolorbox}
\end{minipage}
\caption{%
  Representative examples from \datasetname{}, grouped by reasoning
  complexity.
  \textbf{Top row}: single-hop questions (one spatial relation).
  \textbf{Bottom row}: multi-hop questions chaining two relations
  (\texttt{left\_of}$\to$\texttt{above};
  \texttt{left\_of}$\to$\texttt{left\_of}).
  Underlined phrases mark spatial relations; reasoning is truncated
  for space.
  Full chains are in the released dataset.
}
\label{fig:dataset-examples}
\vspace{-20pt}
\end{figure*}

Figure~\ref{fig:dataset-examples} shows representative examples from the released dataset, grouped by reasoning complexity.
\FloatBarrier
\FloatBarrier

\section{LLM-as-Judge Prompt}
\label{app:judge-prompt}
 
The following prompt is used verbatim for all judge evaluations described in \S\ref{sec:eval:judge}.
The judge model receives the five fields listed below and returns a structured JSON object; no additional instructions are provided at inference time.
 
\paragraph{Input fields.}
\begin{itemize}[leftmargin=*, noitemsep]
  \item \texttt{question}: the spatial VQA question.
  \item \texttt{ground\_truth}: the ground-truth answer derived from the GQA scene graph.
  \item \texttt{scene\_graph}: the relevant scene-graph subgraph (object nodes, attributes, and spatial relation edges).
  \item \texttt{candidate\_answer}: the model's final answer.
  \item \texttt{candidate\_reasoning}: the model's full thinking trace,
        with \texttt{<think>}/\texttt{</think>} delimiters stripped.
\end{itemize}
 
 
\begin{tcolorbox}[
  colback=gray!4, colframe=gray!35,
  title={Spatial Reasoning Judge — System Prompt},
  fonttitle=\small\bfseries,
  left=6pt, right=6pt, top=4pt, bottom=4pt]
  
\small\ttfamily
You are a strict evaluator for spatial reasoning outputs.
 
\medskip
\textbf{Goal}
 
Evaluate a candidate answer and its reasoning using:
\begin{itemize}[leftmargin=1.5em, noitemsep, topsep=2pt]
  \item question
  \item ground truth (when available)
  \item scene graph
  \item candidate final answer
  \item candidate reasoning
  \item image scene graph 
\end{itemize}
 
\medskip
\textbf{Scoring Rubric (0.0 to 1.0)}
 
\smallskip
\underline{1) answer\_correctness}\\
1.0: final answer is correct and matches the expected target.\\
0.5: partially correct or ambiguous but defensible.\\
0.0: clearly incorrect or unsupported.
 
\smallskip
\underline{2) reasoning\_faithfulness}\\
1.0: reasoning is grounded in spatial evidence (relations, coordinates,
object references).\\
0.5: mostly grounded but with weak/missing support in some step.\\
0.0: hallucinated or contradicts available evidence.
 
\smallskip
\underline{3) reasoning\_completeness}\\
1.0: reasoning covers all key steps needed to justify the answer.\\
0.5: key idea present but skips important steps or constraints.\\
0.0: too shallow, irrelevant, or missing.
 
\medskip
\textbf{Output Contract}
 
Return \textbf{only} valid JSON with this exact schema:
 
\medskip
\{\\
\quad "answer\_correctness": 0.0,\\
\quad "reasoning\_faithfulness": 0.0,\\
\quad "reasoning\_completeness": 0.0,\\
\}
 
\medskip
\textbf{Output Constraints}
\begin{itemize}[leftmargin=1.5em, noitemsep, topsep=2pt]
  \item No markdown fences in the final response.
  \item No extra keys.
\end{itemize}

Question: (\texttt{question})

GroundTruth: (\texttt{ground\_truth})

SceneGraph: (\texttt{scene\_graph})

CandidateAnswer: (\texttt{candidate\_answer})

CandidateReasoning (\texttt{candidate\_reasoning})

Return only Json
  
\end{tcolorbox}

\section{Hyperparameter Details}
\label{app:hyperparams}

Tables~\ref{tab:hyperparams-model} and~\ref{tab:hyperparams-training} report the full configuration used to fine-tune Qwen3-VL-8B-Thinking on \textsc{SpatialChain}. The same configuration is used for the 4B variant, with only the base model identifier changed.

\begin{table}[h]
  \centering
  \caption{Model and LoRA configuration.}
  \label{tab:hyperparams-model}
  \begin{tabular}{ll}
    \toprule
    \textbf{Hyperparameter} & \textbf{Value} \\
    \midrule
    Base model                  & Qwen3-VL-8B-Thinking \\
    Quantization                & 4-bit NF4 (BitsAndBytes) \\
    Quantization compute dtype  & bfloat16 \\
    Double quantization         & \checkmark \\
    Attention implementation    & Flash Attention 2 \\
    \midrule
    LoRA rank $r$               & 16 \\
    LoRA $\alpha$               & 32 \\
    LoRA dropout                & 0.05 \\
    RSLoRA                      & \checkmark \\
    Target modules              & all linear layers (excl.\ \texttt{lm\_head}) \\
    \bottomrule
  \end{tabular}
\end{table}

\begin{table}[h]
  \centering
  \caption{Training configuration for SFT on \textsc{SpatialChain}.}
  \label{tab:hyperparams-training}
  \begin{tabular}{ll}
    \toprule
    \textbf{Hyperparameter} & \textbf{Value} \\
    \midrule
    Training examples           & 28,350 \\
    Epochs                      & 2 \\
    Per-device batch size       & 6 \\
    Gradient accumulation steps & 2 \\
    Effective batch size        & 12 \\
    \midrule
    Learning rate               & $3 \times 10^{-5}$ \\
    LR schedule                 & cosine \\
    Warmup ratio                & 0.05 \\
    Weight decay                & 0.01 \\
    Gradient clipping           & 1.0 \\
    Optimiser                   & AdamW (fused) \\
    \midrule
    Precision                   & bfloat16 + TF32 \\
    Gradient checkpointing      & disabled \\
    Max sequence length         & 32,768 tokens \\
    Image max size              & 640 px (longest side) \\
    \midrule
    Hardware                    & 1 $\times$ A100 80\,GB \\
    Training framework          & HuggingFace Transformers + PEFT \\
    Random seed                 & 3407 \\
    \bottomrule
  \end{tabular}
\end{table}

\section{Inference System Prompt}
\label{app:inference-prompt}

The following system prompt is used verbatim for all open-weight
and proprietary models evaluated in \S\ref{sec:exp:setup}.
For InternVL3.5 models, this prompt is prepended with the standard
thinking-activation prefix required by the model family.

\begin{tcolorbox}[
  colback=gray!5, colframe=gray!40,
  title={\textbf{System prompt — inference}},
  fonttitle=\small\bfseries,
  left=6pt, right=6pt, top=4pt, bottom=4pt
]
\small
\texttt{Your task:}\\
\texttt{1. Analyze the image carefully.}\\
\texttt{2. Provide concise reasoning grounded in visible evidence}\\
\phantom{\texttt{1. }}\texttt{from the image.}\\
\texttt{3. End your response with `Answer: <one short sentence>'.}
\end{tcolorbox}

\noindent
For EmbSpatial-Bench evaluations (\S\ref{sec:res:tradeoff}),
the same prompt is used without modification, allowing any
stylistic shift in model outputs to be attributed to fine-tuning
rather than prompt differences. For ERQA there is no system prompt because the task instruction is embedded in each question.

\section{Additional Experimental Results}
\label{app:extra-results}

\subsection{Extended Chain-Quality Metrics}

Table~\ref{tab:app-full-metrics} extends the main comparison
(Table~\ref{tab:model_comparison}) with four additional metrics:
ROUGE-2, BLEU-1, token-level precision and recall reported
separately, entity overlap (fraction of scene-graph object labels
present in the predicted chain), and step markers (mean count of
explicit step-delimiter phrases per chain, ``first'', ''second'', ''third'',
``then'', ''next'',``finally'').

Two patterns emerge beyond what the main table captures.
Token precision vs.\ recall separates models by how they
fail: proprietary models (Claude~4.6, GPT-5.4, InternVL3.5-4B)
show high precision but very low recall, confirming that their
compact outputs contain \emph{relevant} words but cover only a
small fraction of the reference content.
Open-weight base models (Qwen3-VL-4B/8B) exhibit the reverse —
moderate precision but high recall — consistent with their
verbose, exploratory chains.
Step markers provide a direct proxy for explicit
step-by-step structure: fine-tuned models average 0.26–0.45
per chain (matching the reference format), while open-weight base
models average 2–3 (enumerating more steps than the reference),
and proprietary models score 0.00, indicating the near-total
absence of explicit reasoning structure.
The entity overlap column further confirms that fine-tuned models
($\approx 0.48$) reference scene-graph objects more faithfully
than any other group, despite generating shorter chains.

\begin{table}[h]
\centering
\caption{Evaluation of reasoning chains on the \datasetname{}
  test set. Extended version of Table~\ref{tab:model_comparison}
  with additional metrics (ROUGE-2, BLEU, token precision/recall,
  entity overlap, step markers per chain).
  Models sorted by overall accuracy.}
\label{tab:app-full-metrics}
\resizebox{\textwidth}{!}{
\begin{tabular}{lcccccccccc}
\toprule
Model & Acc & R-1 & R-2 & R-L
  & F1\textsubscript{tok} & Prec\textsubscript{tok} & Rec\textsubscript{tok}
  & BLEU-1 & Entity-ov. & Step-mk. \\
\midrule
Qwen3-VL-4B FT  & 82.23 & 0.657 & 0.407 & 0.450 & 0.646 & 0.701 & 0.614 & 0.570 & 0.481 & 0.26 \\
Qwen3-VL-8B FT  & 85.03 & 0.654 & 0.402 & 0.443 & 0.643 & 0.694 & 0.615 & 0.568 & 0.484 & 0.45 \\
GLM-4.1-9B      & 81.34 & 0.438 & 0.212 & 0.301 & 0.421 & 0.474 & 0.470 & 0.302 & 0.305 & 1.34 \\
Qwen3-VL-8B     & 78.88 & 0.425 & 0.218 & 0.293 & 0.412 & 0.353 & 0.597 & 0.328 & 0.368 & 2.15 \\
Qwen3-VL-4B     & 78.44 & 0.403 & 0.208 & 0.279 & 0.392 & 0.323 & 0.613 & 0.303 & 0.379 & 2.77 \\
Claude~4.6      & 79.89 & 0.359 & 0.187 & 0.258 & 0.354 & 0.683 & 0.248 & 0.127 & 0.253 & 0.00 \\
InternVL3.5-8B  & 79.33 & 0.323 & 0.180 & 0.228 & 0.316 & 0.454 & 0.451 & 0.183 & 0.310 & 1.41 \\
InternVL3.5-4B  & 81.34 & 0.314 & 0.175 & 0.227 & 0.305 & 0.658 & 0.271 & 0.158 & 0.210 & 0.26 \\
GPT-5.4         & 72.18 & 0.296 & 0.140 & 0.225 & 0.294 & 0.731 & 0.187 & 0.049 & 0.222 & 0.00 \\
\bottomrule
\end{tabular}}
\end{table}

\subsection{Per-Relation Accuracy}

Table~\ref{tab:app-per-relation} reports per-relation accuracy on the
\datasetname{} test split.
Several relations (\texttt{behind}: 3 examples;
\texttt{below}: 4; \texttt{above}: 12; \texttt{next\_to}: 8)
have very small sample sizes due to the complexity-biased test
selection (\S\ref{sec:dataset:selection}), and results for those
should be interpreted with caution.

For the two well-sampled projective relations
(\texttt{left\_of}: $n=314$; \texttt{right\_of}: $n=327$),
the ranking of models tracks overall accuracy closely,
confirming that the aggregate score is dominated by these
two categories.
The most notable departure is Claude~4.6, which achieves
substantially higher accuracy on non-projective relations
(\texttt{in\_front\_of}: 0.93, \texttt{behind}: 1.00,
\texttt{next\_to}: 1.00) than on projective ones
(\texttt{left\_of}: 0.76, \texttt{right\_of}: 0.83).
This pattern is consistent with Claude~4.6's high shortcut rate
(\S\ref{sec:res:judge}): projective left/right distinctions require
precise visual grounding that shortcuts cannot supply, whereas
relations such as \texttt{in\_front\_of} are more recoverable from
linguistic co-occurrence priors.

\begin{table}
\centering
\caption{Per-relation accuracy on the \datasetname{} test split.
  Counts of validation examples for each relation:
  left\_of (314), right\_of (327), in\_front\_of (15), behind (3),
  above (12), below (4), next\_to (8), other (13).
  Note that some relations have very small sample sizes;
  results for those are reported for completeness only.}
\label{tab:app-per-relation}
\resizebox{\textwidth}{!}{
\begin{tabular}{lcccccccc}
\toprule
Model & left\_of & right\_of & in\_front\_of & behind
  & above & below & next\_to & other \\
\midrule
Qwen3-VL-8B FT  & 0.836 & 0.868 & 0.800 & 1.000 & 0.833 & 0.500 & 0.750 & 0.923 \\
Qwen3-VL-4B FT  & 0.816 & 0.825 & 0.800 & 0.667 & 0.833 & 1.000 & 1.000 & 0.923 \\
GLM-4.1-9B      & 0.812 & 0.808 & 0.933 & 0.667 & 0.833 & 0.500 & 0.750 & 1.000 \\
InternVL3.5-4B  & 0.803 & 0.815 & 1.000 & 0.667 & 0.667 & 1.000 & 0.750 & 1.000 \\
Qwen3-VL-4B     & 0.796 & 0.778 & 0.733 & 0.667 & 0.833 & 0.000 & 0.750 & 0.769 \\
Qwen3-VL-8B     & 0.789 & 0.788 & 0.667 & 1.000 & 0.833 & 0.500 & 0.750 & 0.923 \\
InternVL3.5-8B  & 0.769 & 0.805 & 0.933 & 1.000 & 1.000 & 1.000 & 1.000 & 0.846 \\
Claude~4.6      & 0.763 & 0.828 & 0.933 & 1.000 & 0.833 & 1.000 & 1.000 & 0.846 \\
GPT-5.4         & 0.690 & 0.748 & 0.867 & 1.000 & 0.833 & 0.500 & 0.750 & 0.769 \\
\bottomrule
\end{tabular}}
\end{table}

\subsection{Fine-Tuning Trajectory on External Benchmarks}

Tables~\ref{tab:erqa_checkpoints} and~\ref{tab:embspatial_checkpoints}
track Qwen3-VL-4B accuracy across six checkpoints during SFT on
\datasetname{}: after 63, 313, 625, 1{,}875, and 2{,}813 steps
(approximately 1k, 5k, 10k, 15k samples seen and one full epoch),
plus the final model at 2 epochs (final).
Both external benchmarks are evaluated with the same inference
setup as the main experiments.

\paragraph{ERQA.}
Degradation on ERQA is immediate and sharp: overall accuracy
drops from 0.453 (base) to 0.388 at step 63, before stabilising
between 0.372 and 0.425 for the remainder of training
(Table~\ref{tab:erqa_checkpoints}).
The sharpest per-category drop occurs in
\emph{Trajectory Reasoning} (0.409 $\to$ 0.273 at final),
a category that requires tracking object paths across multiple
images — a capability \datasetname{} does not train on.
\emph{Action Reasoning} and \emph{State} categories show more
moderate and partially recovered degradation ($\approx -0.02$\,pp
from base to final), suggesting these competences are more
resilient to stylistic specialisation.
Notably, the degradation does not worsen monotonically: step~313
recovers partially before declining again, consistent with the
model oscillating between the \datasetname{} format and its
pre-SFT behaviour during early training.

\paragraph{EmbSpatial-Bench.}
The EmbSpatial-Bench trajectory (Table~\ref{tab:embspatial_checkpoints})
shows a more structured pattern.
Projective relations (\texttt{left}, \texttt{right}) and
\texttt{above} are remarkably stable throughout training
(drop $<2$\,pp at final), while depth-related relations
(\texttt{close}: $-5.0$\,pp; \texttt{far}: $-7.4$\,pp)
bear the bulk of the degradation.
This is consistent with the interpretation in
\S\ref{sec:res:tradeoff}: \datasetname{} trains primarily on
projective and containment relations, leaving depth-ordering
competence undertrained in the fine-tuned format.
A partial recovery is visible from step~1{,}875 onward for
\texttt{close} (0.613 $\to$ 0.649 at final), suggesting the
model does not catastrophically forget depth reasoning but
applies the \datasetname{} template inappropriately to
EmbSpatial prompts.

\begin{table}
\centering
\caption{Qwen3-VL-4B accuracy on ERQA by category over fine-tuning.
  Step counts refer to optimisation steps with effective batch size 12.}
\label{tab:erqa_checkpoints}
\resizebox{\textwidth}{!}{
\begin{tabular}{lccccccccc}
\toprule
Checkpoint & Traj. & Action & Spatial & State & Task & Multi-v.
  & Point. & Other & Overall \\
\midrule
Base       & 0.409 & 0.431 & 0.524 & 0.473 & 0.500 & 0.405 & 0.382 & 0.429 & \textbf{0.453} \\
Step 63    & 0.288 & 0.375 & 0.488 & 0.527 & 0.368 & 0.297 & 0.353 & 0.143 & 0.388 \\
Step 313   & 0.424 & 0.458 & 0.464 & 0.418 & 0.500 & 0.324 & 0.441 & 0.071 & 0.425 \\
Step 625   & 0.167 & 0.417 & 0.464 & 0.455 & 0.474 & 0.324 & 0.412 & 0.214 & 0.380 \\
Step 1875  & 0.303 & 0.417 & 0.405 & 0.436 & 0.421 & 0.351 & 0.324 & 0.429 & 0.385 \\
Step 2813  & 0.303 & 0.389 & 0.405 & 0.436 & 0.342 & 0.351 & 0.441 & 0.143 & 0.372 \\
Final      & 0.273 & 0.417 & 0.452 & 0.455 & 0.368 & 0.297 & 0.412 & 0.143 & 0.380 \\
\bottomrule
\end{tabular}}
\end{table}

\begin{table}
\centering
\caption{Qwen3-VL-4B accuracy on EmbSpatial-Bench by relation category
  over fine-tuning.
  Vertical (\emph{above}) and projective (\emph{left}, \emph{right})
  relations are preserved; depth-related relations
  (\emph{close}, \emph{far}) are most affected.}
\label{tab:embspatial_checkpoints}
\begin{tabular}{lccccccc}
\toprule
Checkpoint & Left & Right & Above & Under & Close & Far & Overall \\
\midrule
Base       & 0.943 & 0.905 & 0.782 & 0.772 & 0.699 & 0.640 & \textbf{0.791} \\
Step 63    & 0.937 & 0.913 & 0.797 & 0.698 & 0.676 & 0.608 & 0.773 \\
Step 313   & 0.929 & 0.873 & 0.819 & 0.701 & 0.670 & 0.582 & 0.763 \\
Step 625   & 0.904 & 0.845 & 0.809 & 0.724 & 0.642 & 0.609 & 0.757 \\
Step 1875  & 0.904 & 0.858 & 0.809 & 0.704 & 0.613 & 0.567 & 0.744 \\
Step 2813  & 0.909 & 0.873 & 0.807 & 0.721 & 0.624 & 0.601 & 0.757 \\
Final      & 0.925 & 0.869 & 0.804 & 0.721 & 0.649 & 0.566 & 0.757 \\
\bottomrule
\end{tabular}
\end{table}

\section{Qualitative Example}
\label{app:quali-fig}
Figure~\ref{fig:qualitative-example} contrasts the two most different reasoning regimes on a single test item; complete outputs for all nine models follow in Appendix~\ref{app:extra-quali-samples}.

\begin{figure}[h!]
\centering
\small
\setlength{\fboxsep}{6pt}

\begin{tcolorbox}[
  colback=gray!8, colframe=gray!40,
  title={\textbf{Question} \quad \textit{imageId: 2328879}},
  fonttitle=\small\bfseries, left=4pt, right=4pt, top=3pt, bottom=3pt
]
\textit{``Do you see kittens to the right of the bread that is to
the right of the knife?''}\\[4pt]
\textbf{Ground-truth answer:} No, there is a bird to the right of
the bread.\\[2pt]
\textbf{Reference reasoning (SpatialChain):}
Looking at the image step-by-step:
(1)~knife on the left of the plate;
(2)~bread to the right of the knife;
(3)~gray birds to the right of that bread;
(4)~no kittens present.
\textit{Conclude: No, there is a bird.}
\end{tcolorbox}

\vspace{4pt}

\begin{minipage}[t]{0.48\textwidth}
\begin{tcolorbox}[
  colback=green!6, colframe=green!50,
  title={\textbf{Qwen3-VL-4B FT} — grounded},
  fonttitle=\small\bfseries,
  left=4pt, right=4pt, top=3pt, bottom=3pt
]
\textit{``[\ldots] knife on the left [\ldots] bread to the right of
the knife [\ldots] a bird perched to the right of that bread
[\ldots] no kittens anywhere.''}

\vspace{6pt}
\textbf{Answer:} \textcolor{darkgreen}{\checkmark}~No, there is a bird.

\vspace{4pt}
{\footnotesize\textcolor{gray}{Follows the reference chain
step-by-step; names the bird as direct evidence.}}
\end{tcolorbox}
\end{minipage}
\hfill
\begin{minipage}[t]{0.48\textwidth}
\begin{tcolorbox}[
  colback=blue!5, colframe=blue!35,
  title={\textbf{InternVL3.5-8B} — verbose-decorative},
  fonttitle=\small\bfseries,
  left=4pt, right=4pt, top=3pt, bottom=3pt
]
\textit{``[\ldots] there is indeed a bird, not a kitten [\ldots]
The setting seems to be outdoors, possibly near a body of water
[\ldots] outdoor dining [\ldots]''}

\vspace{6pt}
\textbf{Answer:} \textcolor{darkgreen}{\checkmark}~No (implicit)

\vspace{4pt}
{\footnotesize\textcolor{gray}{Correct answer embedded in scene description unrelated to the spatial chain.}}
\end{tcolorbox}
\end{minipage}

\caption{%
  Both models answer correctly, but through qualitatively different
  processes.
  The reference chain (top) was generated from the GQA scene graph
  by \datasetname{} and follows a step-by-step spatial structure.
  The fine-tuned model (left) closely mirrors this structure and
  names the bird as direct evidence.
  InternVL3.5-8B (right) arrives at the correct answer via hallucinated scene description. Full outputs for all nine models are in
  Appendix~\ref{app:extra-quali-samples}.
}
\label{fig:qualitative-example}
\end{figure}

\section{Additional Qualitative Example}
\label{app:extra-quali-samples}

Table~\ref{fig:qualitative-example} present additional reasoning chain example from the \datasetname{} test split, covering all nine evaluated models and including the GLM-4.1-9B case discussed in \S\ref{sec:res:judge}.

\begin{figure}[h]
    \centering
    \includegraphics[width=0.7\linewidth]{2328879.jpg}
    \caption{Image 2328879 used as input fot the discussed question "Do you see kittens to the right of the bread that is to the right of the knife".}
    \label{fig:image-2328879}
\end{figure}


\begin{longtable}{@{}p{0.10\textwidth} p{0.70\textwidth} c@{}}
\caption{%
  Complete model outputs for the qualitative example in
  Figure~\ref{fig:qualitative-example}
  (\textit{imageId}: 2328879).
  \textbf{Question:} ``Do you see kittens to the right of the bread
  that is to the right of the knife?''
  \textbf{Ground-truth answer:} No, there is a bird to the right of the bread.
  Outputs are reproduced verbatim; \texttt{<think>}\,/\,\texttt{</think>}
  delimiters are omitted for readability.
}
\label{tab:app-qualitative-full}\\
\toprule
\textbf{Model} & \textbf{Full prediction} & \textbf{Correct} \\
\midrule
\endfirsthead
 
\multicolumn{3}{@{}l}{\small\textit{Table~\ref{tab:app-qualitative-full} continued}}\\
\toprule
\textbf{Model} & \textbf{Full prediction} & \textbf{Correct} \\
\midrule
\endhead
 
\midrule
\multicolumn{3}{r@{}}{\small\textit{Continued on next page}}\\
\endfoot
 
\bottomrule
\endlastfoot
 
\rowcolor{gray!10}
\textbf{Reference} \newline
(scene graph)
&
\textit{%
``Looking at the image, let me trace through this step-by-step:
(1) Locating the knife: I can see a knife on the left side of the plate,
positioned near the bottom left.
(2) Finding the bread to the right of the knife: there is a large piece
of bread that sits on the plate to the right of where the knife is placed.
(3) Examining what is to the right of that bread: to the right of the
bread I can see what appears to be gray birds standing on the plate.
These are birds with wings and legs visible.
(4) Looking for kittens: I do not see any kittens anywhere in the image.
The animals present in the scene are birds — specifically gray colored
birds that appear to be standing on or near the bread on the plate.''}
\newline\newline
\textbf{Answer:} No, there is a bird to the right of the bread.
&
— \\[6pt]
 
\rowcolor{green!5}
\textbf{Qwen3-VL-4B FT} \newline
\textit{(fine-tuned)}
&
\textit{%
``Looking at the image, I need to trace through this step by step:
(1) First, I can see a knife on the left side of the image, positioned
near the edge.
(2) To the right of the knife, there is bread visible in the center-left
portion of the image.
(3) To the right of that bread, I can see a white plate with a bird
perched on it.
(4) Looking further to the right of the bread (and the plate with the
bird), I see what appears to be a small bird on the table, but examining
the entire right side of the bread, I do not see any kittens anywhere in
the image.
The image shows a table setting with a woman, food items including bread,
a plate, a bird, and other dining elements.
There are no kittens visible anywhere in the scene, let alone to the
right of the bread.''}
\newline\newline
\textbf{Answer:} No, there is a bird to the right of the bread.
&
\textcolor{green!60!black}{\checkmark} \\[6pt]
 
\rowcolor{green!5}
\textbf{Qwen3-VL-8B FT} \newline
\textit{(fine-tuned)}
&
\textit{%
``Looking at the image, I can see a white plate with various food items on it. On the left side, there's a knife, and to the right of it is a piece of bread. Moving further to the right of the bread, I can see what appears to be a small brown bird standing on the plate.
However, I don't see any kittens anywhere in this image. The only animal visible to the right of the bread is a small brown bird. There are no kittens present in the scene.''}
\newline\newline
\textbf{Answer:} No, there are no kittens visible in the image.
&
\textcolor{green!60!black}{\checkmark} \\[6pt]
 
\rowcolor{orange!5}
\textbf{Qwen3-VL-4B} \newline
\textit{(base)}
&
\textit{%
``So, let us break this down. First, the question is asking if there are kittens to the right of the bread that is to the right of the knife. Let us identify each element. First, the knife: there is a knife on the left side of the plate, near the bread. The bread is the sandwich in the center of the plate. Now, `the bread that is to the right of the knife' — so the knife is on the left, the bread is to the right of that knife. Then, we need to look to the right of that bread. What is there? On the plate, to the right of the bread (sandwich), there is a sauce (red sauce, maybe tomato) and some other food. But the question is about kittens. Are there any kittens in the image? Let us check the image. The image shows a plate with a sandwich, some sauce, a knife, and two small birds (sparrows) on the plate or near it. There are no kittens visible. The birds are sparrows, not kittens.
So, to the right of the bread (which is the sandwich), there are no kittens. The birds are there, but they are birds, not kittens. So the answer is no.''}
\newline\newline
\textbf{Answer:} No.
&
\textcolor{green!60!black}{\checkmark} \\[6pt]
 
\rowcolor{orange!5}
\textbf{Qwen3-VL-8B} \newline
\textit{(base)}
&
\textit{%
``Got it, let's break this down. First, identify the elements. The question is about kittens to the right of the bread that's to the right of the knife. Let's look at the image. First, find the knife. There's a knife on the left side of the plate, next to the fork. Then, the bread: the hamburger bun is on the plate, to the right of the knife. Now, to the right of that bread (hamburger bun), what's there? There's a bird, not a kitten. The image shows birds, not kittens. So no kittens are present. The question asks if there are kittens to the right of the bread (hamburger) which is right of the knife. Since there are birds, not kittens, the answer is no.''}
\newline\newline
\textbf{Answer:}The image shows birds (not kittens) to the right of the hamburger bun, which is to the right of the knife. No kittens are visible. Answer: No, there are no kittens in the image
&
\textcolor{green!60!black}{\checkmark} \\[6pt]
 
\rowcolor{yellow!5}
\textbf{GLM-4.1-9B}
&
\textit{%
``Got it, let's analyze the image. First, look for the knife, then the bread to its right, then check if there are kittens to the right of that bread. The image shows a plate with food, a knife on the left, bread (the bun) next to it, and then birds, not kittens. So there are no kittens in that area. Therefore, the answer would be no.''}
\newline\newline
\textbf{Answer:} To determine this, first locate the knife on the left side of the plate, then identify the bread (bun) to the right of the knife. Next, check for kittens to the right of that bread. The area to the right of the bread shows birds, not kittens. Therefore, the answer is No.
&
\textcolor{green!60!black}{\checkmark} \\[6pt]
 
\rowcolor{blue!4}
\textbf{InternVL3-4B}
&
\textit{%
``kittens is not to the right of the bread that is to the right of the knife.''}
\newline\newline
\textbf{Answer:} No.
&
\textcolor{green!60!black}{\checkmark} \\[6pt]
 
\rowcolor{blue!4}
\textbf{InternVL3-8B}
&
\textit{%
``In the image, to the right of the bread that is to the right of the knife, there is indeed a bird, not a kitten. The bird appears to be a sparrow, standing on the plate beside the plate with the sandwich. There are no kittens visible in the image, so the presence of a kitten to the right of the mentioned bread is not accurate based on what can be observed. The bird is interacting with the food, which adds a dynamic element to the scene, contrasting with the stillness of the sandwich and the remnants of sauce on the plate. The setting seems to be outdoors, possibly near a body of water, given the presence of what appears to be a waterfront in the background.
The interaction between wildlife and human food can sometimes occur in outdoor dining settings, especially in areas where birds are accustomed to human presence.''}
\newline\newline
\textbf{Answer:} No (implicit).
&
\textcolor{green!60!black}{\checkmark} \\[6pt]
 
\rowcolor{purple!4}
\textbf{Claude~4.6}
&
\textit{%
``To the right of the bread/sandwich on the plate, there are birds (sparrows) perched on the edge of the plate, not kittens. No cats or kittens are visible anywhere in the image.''}
\newline\newline
\textbf{Answer:} No.
&
\textcolor{green!60!black}{\checkmark} \\[6pt]
 
\rowcolor{purple!4}
\textbf{GPT-5.4}
&
\textit{%
``The knife is on the left side of the plate, and the bread is to its right. To the right of the bread there are two birds on the plate, not kittens.''}
\newline\newline
\textbf{Answer:} No.
&
\textcolor{green!60!black}{\checkmark} \\
 
\end{longtable}

\section{Judge Validation and Robustness}
\label{app:judge-validation}

This appendix reports the experiments that validate the faithfulness
axis: a human agreement study, a second judge from a different
provider, a sweep of the faithfulness threshold, formal cross-model
correlation coefficients, and a stratification by question complexity.

\subsection{Human Validation of the Judge}
\label{app:human-validation}

We sampled 198 test items — 22 from each of the nine evaluated
models — stratified across the judge's faithfulness range, with
deliberate oversampling around $\mathit{RF}=0.5$. Six annotators
scored every item under the Appendix~\ref{app:judge-prompt} rubric
verbatim, seeing the image, question, ground truth, candidate answer
and chain, with model identity and judge scores hidden. Every item
was scored by two annotators independently (396 annotations).
Annotators scored against the \emph{image}; the judge scores against
the scene graph.

We report Spearman rank correlation on the continuous scores without
binarising either side, since thresholding one rater against another
conflates disagreement about a chain with disagreement about where a
cutoff belongs (Table~\ref{tab:human-agreement}). The
\emph{judge $\leftrightarrow$ individual human} column is the
like-for-like comparison with the human--human protocol.

\begin{table}[h]
  \centering
  \caption{Rank correlation (Spearman $\rho$) between raters on
    $n=198$ items. Mean absolute error of the judge against the
    human consensus is 0.14 (correctness), 0.27 (faithfulness),
    0.21 (completeness).}
  \label{tab:human-agreement}
  \small
  \begin{tabular}{lccc}
    \toprule
    Dimension & human $\leftrightarrow$ human
      & judge $\leftrightarrow$ individual human
      & judge $\leftrightarrow$ human consensus \\
    \midrule
    Answer correctness      & 0.673 & 0.655 & 0.679 \\
    Reasoning faithfulness  & 0.443 & 0.488 & 0.574 \\
    Reasoning completeness  & 0.416 & 0.551 & 0.643 \\
    \bottomrule
  \end{tabular}
\end{table}

On both reasoning dimensions the judge agrees with a human annotator
at least as well as two annotators agree with each other
($\mathit{RF}$: 0.488 vs.\ 0.443; $\mathit{RC}$: 0.551 vs.\ 0.416).
Faithfulness is an intrinsically subjective judgment, and
human--human agreement sets the ceiling one can expect of any rater;
the judge operates at that ceiling. Annotator variance is substantial
and is documented with the released annotations.

\paragraph{What the $\mathit{RF}\leq 0.5$ cutoff captures.}
Among correct answers, a recurring low-faithfulness pattern is a
chain that restates the question and asserts the answer, introducing
no vocabulary beyond the question, identifying no object and
traversing no relation — e.g.\ \textit{``Is there a teddy bear to the
right of the bears that are hanging from the ceiling?''} $\to$
\textit{``Yes, there is a teddy bear to the right of the bears that
are hanging from the ceiling.''} (judge $\mathit{RF}$ 0.5; human mean
0.30). Fourteen chains in the sample follow this template; both the
judge and the annotators score them low. The cutoff also catches the
opposite extreme: a 13{,}000-word Qwen3-VL-8B chain that cycles
through reversals without resolving the relation and terminates
mid-deliberation, yet is scored correct (judge $\mathit{RF}$ 0.2;
both annotators 0.0). Length and grounding are orthogonal.

\paragraph{Where judge and annotators diverge.}
In the 56 items where the judge scores faithfulness low and the
annotators score it high, the chains are not padded: relative to
chains the judge scores highly they have the same length (median 104
vs.\ 105 words) and no more scene-description material, but mention
fewer spatial relations (7.7 vs.\ 10.4) and fewer step markers (0.52
vs.\ 0.89). This is an asymmetry of evidence rather than an error by
either rater: the judge penalises a chain that reaches the right
answer without engaging the annotated relational structure, whereas
a human inspecting only the image finds such a chain reasonable.

\subsection{Second Judge from a Different Provider}
\label{app:second-judge}

We re-ran the full judging protocol on all 899 test items with
GPT-5.6, using identical code and prompt. Continuous faithfulness
scores correlate at Spearman $\rho=0.74$ (quadratic $\kappa=0.74$),
pass/fail verdicts agree at $\kappa=0.84$, and the ranking of models
by shortcut rate is preserved ($\rho=0.88$; Table~\ref{tab:second-judge}).
GPT-5.6 is uniformly stricter, which shifts absolute values but not
the comparative conclusions. The same-family model, Claude~Sonnet~4.6,
ranks 5th of nine under the Claude judge and 3rd under the GPT judge —
it is treated more harshly, not more leniently, by the same-family judge.

\begin{table}[h]
  \centering
  \caption{Shortcut rate under two independent judges
    (fraction of correct test set, $\mathit{RF}\leq 0.5$).}
  \label{tab:second-judge}
  \small
  \begin{tabular}{lcc}
    \toprule
    Model & Claude Sonnet 4.6 judge & GPT-5.6 judge \\
    \midrule
    Qwen3-VL-8B FT     & 0.222 & 0.466 \\
    Qwen3-VL-4B FT     & 0.194 & 0.396 \\
    GLM-4.1-9B         & 0.453 & 0.595 \\
    Claude Sonnet 4.6  & 0.390 & 0.546 \\
    Qwen3-VL-4B        & 0.264 & 0.574 \\
    Qwen3-VL-8B        & 0.260 & 0.588 \\
    GPT-5.4            & 0.246 & 0.373 \\
    InternVL3.5-8B     & 0.473 & 0.653 \\
    InternVL3.5-4B     & 0.580 & 0.699 \\
    \bottomrule
  \end{tabular}
\end{table}

\subsection{Threshold Sensitivity}
\label{app:threshold}

\begin{figure}[h]
  \centering
  \includegraphics[width=0.75\linewidth]{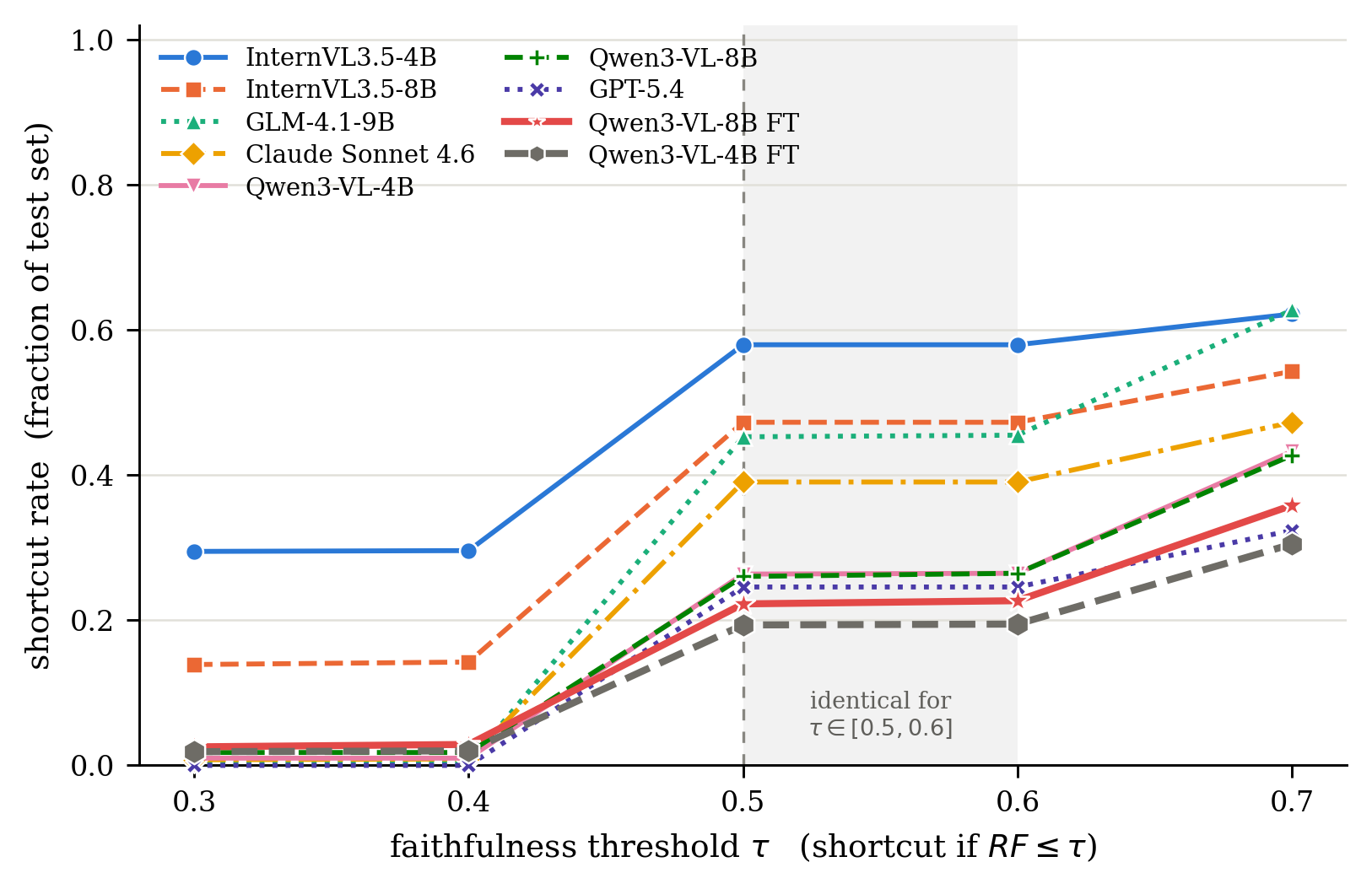}
  \caption{Shortcut rate as a function of the faithfulness threshold
    $\tau$. The shaded region marks the plateau $\tau\in[0.5,0.6]$ on
    which all rates are identical.}
  \label{fig:threshold-sweep}
\end{figure}

We swept the faithfulness cutoff over $\tau\in[0.3,0.7]$ for all nine
models (Table~\ref{tab:threshold-sweep}, Figure~\ref{fig:threshold-sweep}).
Because the judge's scores concentrate on a few discrete values,
shortcut rates are identical for $\tau\in[0.5,0.6]$ and the model
ordering is unchanged across that plateau — the published $0.5$
cutoff sits in the middle of a flat region rather than at a
sensitive point. Below $\tau=0.4$ the metric has essentially no
discriminative power: seven of nine models fall below $0.03$. The
substantive conclusions — that the fine-tuned variants have the
lowest shortcut rates and InternVL3.5-4B/8B the highest — hold
throughout the range where the metric separates models at all
($\tau\geq 0.5$).

\begin{table}[h]
  \centering
  \caption{Shortcut rate vs.\ faithfulness threshold $\tau$
    ($\mathit{RF}\leq\tau$; fraction of correct test set).}
  \label{tab:threshold-sweep}
  \small
  \begin{tabular}{lccccc}
    \toprule
    Model & $\tau{=}0.3$ & $\tau{=}0.4$ & $\tau{=}0.5$ & $\tau{=}0.6$ & $\tau{=}0.7$ \\
    \midrule
    Qwen3-VL-8B FT     & 0.026 & 0.029 & 0.222 & 0.227 & 0.358 \\
    Qwen3-VL-4B FT     & 0.019 & 0.020 & 0.194 & 0.195 & 0.305 \\
    GLM-4.1-9B         & 0.003 & 0.003 & 0.453 & 0.455 & 0.627 \\
    Claude Sonnet 4.6  & 0.008 & 0.009 & 0.390 & 0.390 & 0.473 \\
    Qwen3-VL-4B        & 0.010 & 0.010 & 0.264 & 0.265 & 0.433 \\
    Qwen3-VL-8B        & 0.018 & 0.018 & 0.260 & 0.265 & 0.427 \\
    GPT-5.4            & 0.000 & 0.000 & 0.246 & 0.246 & 0.324 \\
    InternVL3.5-8B     & 0.139 & 0.142 & 0.473 & 0.473 & 0.543 \\
    InternVL3.5-4B     & 0.295 & 0.296 & 0.580 & 0.580 & 0.622 \\
    \bottomrule
  \end{tabular}
\end{table}

\subsection{Cross-Model Correlation between Accuracy and Chain Quality}
\label{app:crossmodel-corr}

Table~\ref{tab:crossmodel-corr} reports formal correlation
coefficients between a model's VQA accuracy and its mean chain
quality (ROUGE-1) across the models of Table~\ref{tab:model_comparison}.
Across all nine models the association is moderate and at best
marginally significant, and it is carried largely by the two
fine-tuned variants, which are simultaneously the most accurate
systems and, by construction, the ones whose chains most closely
match the reference style. With nine models these coefficients are
indicative rather than conclusive. R-L and token-F1 give
near-identical values (all nine models, Acc $\times$
F1\textsubscript{tok}: $r=+0.681$, $\rho=+0.644$, $\tau=+0.479$).

\begin{table}[h]
  \centering
  \caption{Cross-model correlation between VQA accuracy and
    reasoning-chain quality (ROUGE-1).}
  \label{tab:crossmodel-corr}
  \small
  \begin{tabular}{lcccc}
    \toprule
    Subset & $n$ & Pearson $r$ & Spearman $\rho$ & Kendall $\tau$ \\
    \midrule
    All models        & 9 & $+0.690$ ($p=0.040$) & $+0.644$ ($p=0.061$) & $+0.479$ ($p=0.075$) \\
    Zero-shot only    & 7 & $+0.445$ ($p=0.317$) & $+0.270$ ($p=0.558$) & $+0.195$ ($p=0.543$) \\
    Qwen3 family only & 4 & $+0.926$ ($p=0.074$) & $+0.800$ ($p=0.200$) & $+0.667$ ($p=0.333$) \\
    \bottomrule
  \end{tabular}
\end{table}

\subsection{Single-Hop vs.\ Multi-Hop Questions}
\label{app:hops}

We stratified the test set by hop count — the number of spatial
relations chained in the question, inferred from the question text
(385 single-hop, 509 multi-hop, 5 ambiguous cases classified by
hand) — and recomputed accuracy and faithfulness per stratum
(Table~\ref{tab:hops}). The effect is small and, for most models,
runs mildly opposite to the intuition that multi-hop is harder:
accuracy averages $+0.019$ higher on multi-hop and faithfulness
$+0.014$ higher, with the Qwen3 base models showing the clearest
positive shift. The two models most associated with shortcuts,
InternVL3.5-4B/8B, are flat or slightly negative. We read this as
evidence that the dominant difficulty is perceiving the individual
relation rather than composing several.

Accuracy here is computed over all items in each stratum using the
judge's correctness label, whereas Table~\ref{tab:model_comparison}
excludes unparsed predictions per model; this accounts for the small
residual difference for InternVL3.5-4B ($+0.018$) and InternVL3.5-8B
($+0.014$), the two models with the most unparsed outputs. For the
other seven models the two agree to within $0.005$.

\begin{table}[h]
  \centering
  \caption{Accuracy and reasoning faithfulness by question complexity
    (single-hop: $n=385$; multi-hop: $n=509$). $\Delta$ is multi $-$ single.}
  \label{tab:hops}
  \small
  \begin{tabular}{lcccccc}
    \toprule
    Model & Acc single & Acc multi & $\Delta$Acc & RF single & RF multi & $\Delta$RF \\
    \midrule
    Qwen3-VL-8B FT     & 0.847 & 0.845 & $-0.002$ & 0.630 & 0.630 & $-0.000$ \\
    Qwen3-VL-4B FT     & 0.834 & 0.809 & $-0.024$ & 0.641 & 0.623 & $-0.017$ \\
    GLM-4.1-9B         & 0.810 & 0.815 & $+0.005$ & 0.517 & 0.533 & $+0.016$ \\
    Claude Sonnet 4.6  & 0.777 & 0.823 & $+0.047$ & 0.529 & 0.550 & $+0.022$ \\
    Qwen3-VL-4B        & 0.764 & 0.805 & $+0.042$ & 0.566 & 0.597 & $+0.031$ \\
    Qwen3-VL-8B        & 0.748 & 0.815 & $+0.067$ & 0.550 & 0.586 & $+0.036$ \\
    GPT-5.4            & 0.694 & 0.741 & $+0.047$ & 0.529 & 0.551 & $+0.022$ \\
    InternVL3.5-8B     & 0.813 & 0.803 & $-0.009$ & 0.451 & 0.436 & $-0.016$ \\
    InternVL3.5-4B     & 0.831 & 0.831 & $-0.000$ & 0.327 & 0.360 & $+0.033$ \\
    \bottomrule
  \end{tabular}
\end{table}



\end{document}